%% file: main.tex
\documentclass[10pt,twocolumn,letterpaper]{article}

\usepackage{iccv}              
\usepackage{graphicx}
\usepackage{tabularray}
\usepackage{array}
\usepackage{multirow}
\UseTblrLibrary{diagbox}

\definecolor{iccvblue}{rgb}{0.21,0.49,0.74}
\usepackage[pagebackref,breaklinks,colorlinks,allcolors=iccvblue]{hyperref}
\def\paperID{10179} 
\def\confName{ICCV}
\def\confYear{2025}

\title{FairHuman: Boosting Hand and Face Quality in Human Image Generation \\with Minimum Potential Delay Fairness in Diffusion Models}

\author{Yuxuan Wang$^{1}$\\
{\tt\small wangyuxuan2019@bupt.edu.cn}
\and
Tianwei Cao$^{1 \dagger}$\\
{\tt\small caotianwei@bupt.edu.cn}
\and
Huayu Zhang$^{2}$\\
{\tt\small zhanghy56@chinatelecom.cn}
\and
Zhongjiang He$^{2}$\\
{\tt\small hezhongj\_1@163.com}
\and
Kongming Liang$^{1 \dagger}$\\
{\tt\small liangkongming@bupt.edu.cn}
\and
Zhanyu Ma$^{1}$\\
{\tt\small mazhanyu@bupt.edu.cn}
\\
\and
$^1$ Beijing University of Posts and Telecommunications
$^2$ Institute of Artificial Intelligence, China Telecom\\
}
\begin{document}
\twocolumn[{%
\renewcommand\twocolumn[1][]{#1}%
\vspace{-3em}
\maketitle
\begin{center}
    \centering
    \captionsetup{type=figure}
    \includegraphics[width=\textwidth]{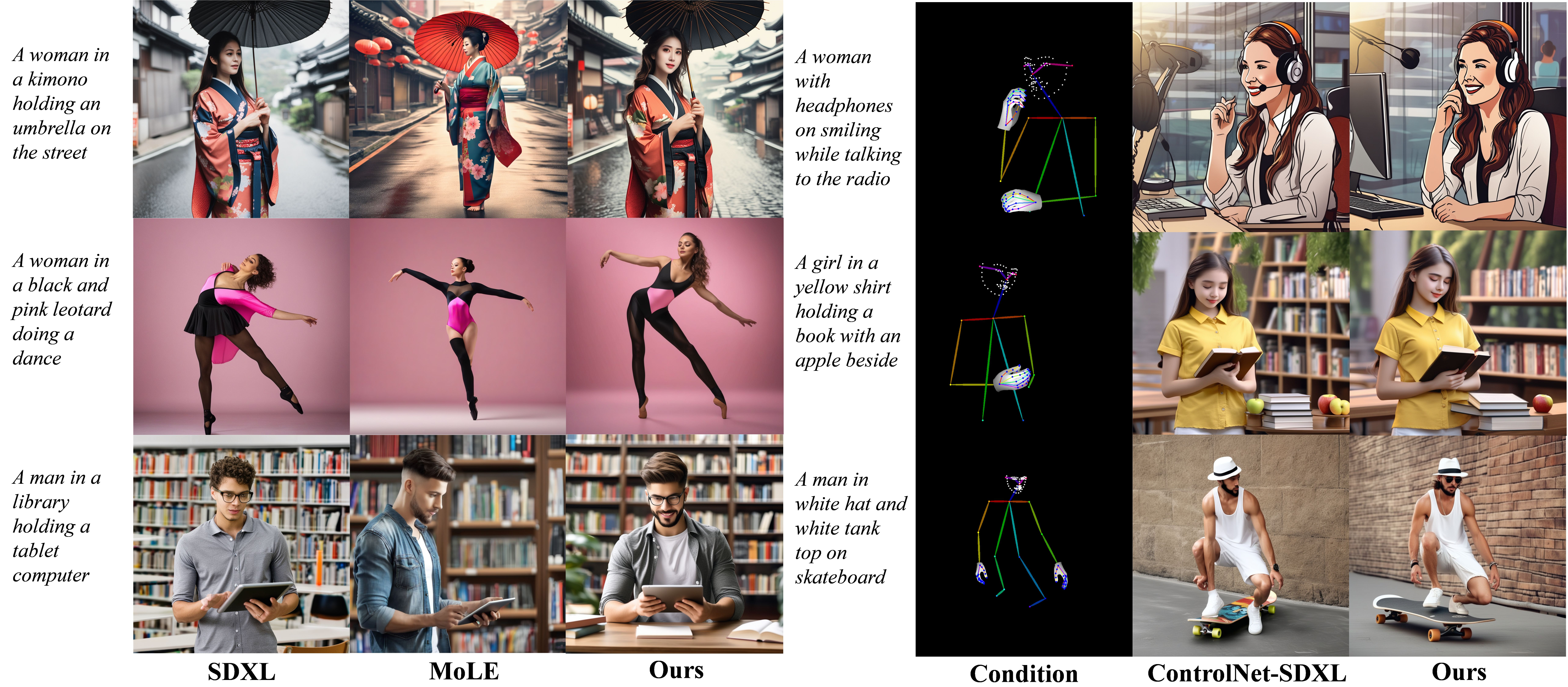}
    \captionof{figure}{Compare with existing methods under two scenarios: \textbf{1) General T2I Generation (in the left part):} from left to right are SDXL-base\cite{podell2023sdxl}, MoLE\cite{zhu2024mole} and Ours. \textbf{2) Controllable Generation (in the right part):} from left to right are input condition, ControlNet-SDXL\cite{zhang2023adding} and Ours. Zoom in for a better view of details.}
    \label{fig:1}
\end{center}%
}]
\let\thefootnote\relax\footnotetext{$\dagger$ Corresponding Author}
\input{sec/0_abstract}    
\input{sec/1_intro}

\input{sec/2_related_work}
\input{sec/3_method}

\input{sec/4_experiments}
\input{sec/5_conclusion}

{
    \small
    \bibliographystyle{ieeenat_fullname}
    \bibliography{main}
}

\input{sec/X_suppl}

\end{document}

%% file: sec/0_abstract.tex
\begin{abstract}
Image generation has achieved remarkable progress with the development of large-scale text-to-image models, especially diffusion-based models. However, generating human images with plausible details, such as faces or hands, remains challenging due to insufficient supervision of local regions during training. To address this issue, we propose FairHuman, a multi-objective fine-tuning approach designed to enhance both global and local generation quality fairly. Specifically, we first construct three learning objectives: a global objective derived from the default diffusion objective function and two local objectives for hands and faces based on pre-annotated positional priors. Subsequently, we derive the optimal parameter updating strategy under the guidance of the Minimum Potential Delay (MPD) criterion, thereby attaining fairness-aware optimization for this multi-objective problem. Based on this, our proposed method can achieve significant improvements in generating challenging local details while maintaining overall quality. Extensive experiments showcase the effectiveness of our method in improving the performance of human image generation under different scenarios. Our code is available at \href{https://github.com/PRIS-CV/FairHuman}{https://github.com/PRIS-CV/FairHuman}.
\end{abstract}

%% file: sec/1_intro.tex
\section{Introduction}
\label{sec:intro}
\begin{figure}[ht]
  \centering
   \includegraphics[width=1.0\linewidth]{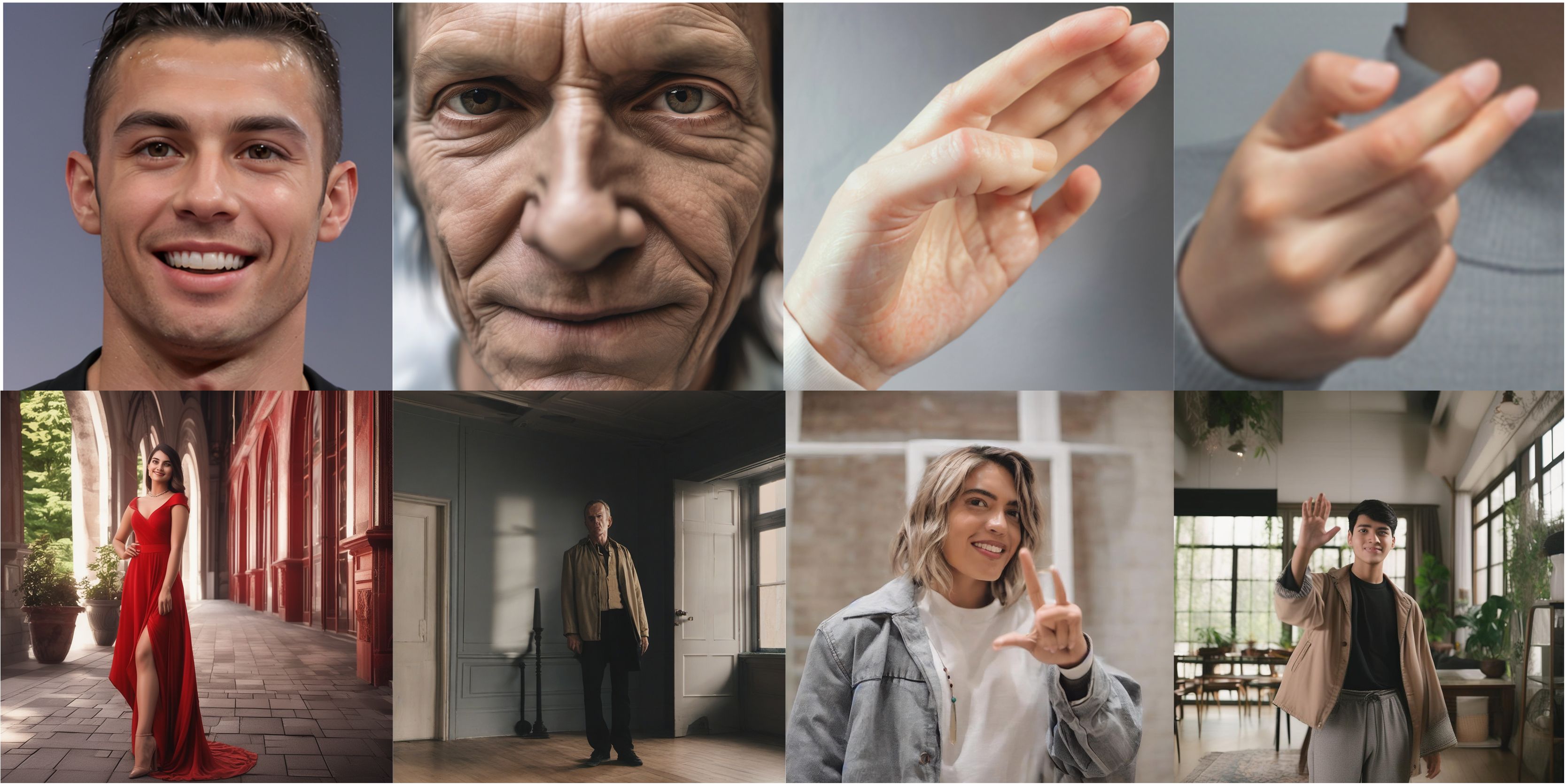}
   \caption{\textbf{Examples of human-related images generated by SDXL.} The first row shows the close-up view images with prompts like ``a close-up human face" and ``a detailed hand gesture". The second row shows the human images under different scenes. Zoom in for a better view of details.}
   \label{fig:3}
\end{figure}
Diffusion-based models have revolutionized image generation by achieving unprecedented high quality and realism through iterative denoising processes. However, their capability of human image generation remains challenged by persistent artifacts in anatomically complex regions, particularly faces and hands. These contents often suffer from generating degradation such as distorted facial expressions and implausible hand poses.

To address this issue, previous work, including ControlNet\cite{zhang2023adding}, HumanSD\cite{ju2023humansd}, and HyperHuman\cite{liu2023hyperhuman}, leverages anatomical information like pose or depth to guide human image generation. Moreover, a leading approach \cite{zhu2024mole} fine-tunes two low-rank modules w.r.t hand and face generation, respectively. Then it uses the soft Mixture of Experts (MoE) mechanism\cite{puigcerver2023sparse} to fuse these modules. Essentially, these methods aim to enhance human image generation by injecting specific knowledge about hands and faces into the generative model.

Despite the success of these methods, the balance between coarse appearance (global quality) and local details (quality of faces and hands) has not been addressed. Through the visual feedback of the human-related images generated by SDXL-base\cite{podell2023sdxl} in Fig.\ref{fig:3}, we have noticed an interesting phenomenon: When the model generates faces or hands separately as the close-up view images, it usually ensures sufficient detail and high quality in terms of structure and texture. However, when generating whole-body images (including both faces and hands), especially when their regions occupy a relatively small portion, noticeable distortions often occur. It suggests that the model inherently possesses the capability to generate plausible faces and hands. Nonetheless, the existence of other contents such as the background, reduces the focus on local regions, thus leading to a decline in generation quality. In addition, a post-processing method called ADetailer\cite{adetailer} repairs the faces and hands by cropping out the distorted regions from the original image and repainting them using super-resolution techniques. Its effectiveness further supports our hypothesis. In this way, the level of attention to global and local aspects is indeed a significant factor influencing the generation quality.  Relevant details will be elaborated in Sec.\ref{sec:problem formulation}.

Motivated by this observation, we decide to construct specific objective functions for global and local generation, respectively. We notice that the default objective function in Eq.(\ref{eq:mse_ori}) for the diffusion model is based on the global mean squared error (MSE), which pays more attention to the overall performance. For complex tasks like human image generation, a single-objective mechanism may lead to the model excelling in certain aspects while underperforming in others. Therefore, here we transform the human image generation task into a multi-objective optimization problem and divide it into three different sub-objectives. Specifically, our optimization objectives include:
\begin{itemize}
    \item \textbf{Facial generation quality:} the naturalness of facial structure and expressions.
    \item \textbf{Hand generation quality:} the plausibility of hand poses and shapes.
    \item \textbf{Global generation quality:} the coherence and perceptual quality of the overall image.
\end{itemize}
As for the construction of objective functions, we retain the default MSE loss function for global features. Then, we create masks using pre-annotated positional information from the dataset and construct mask-based loss functions for local features, namely faces and hands. 

Considering the optimization method, we need to identify an appropriate strategy to maximize the improvement of each objective. Since faces and hands are more difficult to learn compared to global features during training, directly deploying a fixed or equal weight allocation, \ie, the degree of importance, is unreasonable. To achieve the Pareto optimality of human image generation performance, inspired by the idea of fairness criteria utilized in communication networks\cite{mo2000fair}, we incorporate the minimum potential delay (MPD) fairness principle\cite{massoulie1999bandwidth} into our optimization process for parameter updates. Specifically, it dynamically allocates the weight for each objective-specific gradient following the fairness constraints. In this way, the three objective functions mentioned above can be optimized simultaneously within a single diffusion model while mitigating potential conflicts between them. Technically, it can be deployed on fine-tuning different modules like LoRA\cite{hu2021lora} and ControlNet\cite{zhang2023adding} for certain generation tasks.

To this end, we propose a framework called \textbf{FairHuman} to generate human images with higher fidelity and quality under different scenarios. The key insight is that human image generation is a complex task that involves both global and local optimization. The previous single-objective methodology cannot improve the performance of different aspects at the same time. Therefore, we regard the learning for human image generation as a multi-objective optimization process and implement the MPD fairness principle to achieve a more effective and balanced performance. 

Our method contains three procedures (see Fig.\ref{fig:2}): Firstly, we construct separate objective functions for global and regional generation by utilizing the default MSE loss function and regional masks obtained from the pre-annotated positional information. Secondly, we implement the minimum potential delay fairness principle for the joint optimization of all objectives. Eventually, we fine-tune LoRA\cite{hu2021lora} and ControlNet\cite{zhang2023adding} based on the mentioned strategy for two scenarios of human image generation.

In summary, the contributions of our paper are the following:
\begin{itemize}
    \item We emphasize that the poor performance in generating local details of human images stems from insufficient attention during the training process. Based on this factor, we construct different objective functions to focus on global and local generation separately.
    \item We propose FairHuman, a general multi-objective optimization framework for diffusion model in improving the performance of human image generation. Here, we optimize different objective functions simultaneously within a single model based on the minimum potential delay (MPD) fairness principle. To our knowledge, it has not been applied in the diffusion model before.
    \item We implement a widely adopted diffusion model SDXL\cite{podell2023sdxl} as the backbone and deploy our method on different modules, including LoRA\cite{hu2021lora} and ControlNet\cite{zhang2023adding}. Extensive experimental results demonstrate that our method yields a superior performance in enhancing the generation of human images under diverse scenarios.
\end{itemize}

%% file: sec/2_related_work.tex
\section{Related Work}
\label{sec:Related Work}

\textbf{Human Image Generation.} Human image generation refers to generating high-quality and realistic human images complying with the given conditions (\eg, text, image). Among image synthesis tasks, it has garnered significant attention in recent years due to its considerable academic research values and application potentials\cite{jia2024human}. Driven by advancements in scalability and training techniques, existing text-to-image models have demonstrated the ability to generate images with increasing resolution and fidelity. Following the success of GANs\cite{zhang2017stackgan,sauer2023stylegan,zhang2018stackgan++}, diffusion models\cite{ho2020denoising,song2020denoising} has emerged as a powerful alternative and become the predominant architecture currently. Models like Stable Diffusion\cite {rombach2022high} and DALL·E 2\cite{ramesh2022hierarchical} leverage latent diffusion processes to generate highly detailed and semantically accurate images from textual descriptions. However, when creating human-centric images, the latest models\cite{podell2023sdxl,betker2023improving,esser2024scaling} still struggle to generate fine-grained human parts, especially faces and hands. Previous approaches such as\cite{ju2023humansd,liu2023hyperhuman,lu2024handrefiner,wang2024mixture} incorporate supplementary conditions like human pose to achieve better generation in a controllable pattern but have major limitations in consistency. A concurrent work\cite{zhu2024mole} deals with the issue of faces and hands by utilizing two low-rank modules\cite{hu2021lora} with the Mixture of Experts(MoE)\cite{jacobs1991adaptive} mechanism. However, it requires fine-tuning each expert module separately, which brings additional training costs. Meanwhile, stacking of multiple modules also affects the generation efficiency. In our work, we implement joint optimization for both global and local generation performance following the minimum potential delay fairness principle.
\\
%
\textbf{Datasets for Human Image Generation.} The availability of high-quality datasets has significantly propelled the development of human image generation models. These datasets\cite{zheng2015scalable,ionescu2013human3,choi2021viton,ju2023human,fu2022stylegan} provide the necessary diversity and richness required for training models in generating realistic and varied human images. However, most of them have two main drawbacks: \textbf{1)} low resolution and poor image quality, which are inadequate to provide valid prior knowledge for diffusion models. \textbf{2)} lack of annotations on morphological and spatial information of the human body, which cannot effectively guide the model in generating delicate parts like human faces and hands. To alleviate the above limitations, we collect images from available high-quality human-centric datasets\cite{li2024cosmicman,narasimhaswamy2022whose} that contain high-resolution real-world images. To further guarantee the image quality, we apply detectors\cite{redmon2016you,yang2023effective,pavlakos2024reconstructing} to filter out undesirable images while maintaining as much valuable information as possible. In addition, we annotate the body information for controllable generation and positional information for constructing mask-based objective functions. Implementing the procedures mentioned above, we can achieve enhanced training performance aligned with our objectives.


%% file: sec/3_method.tex
\section{Methodology}\label{sec:Method}
\begin{figure*}[t]
\centering
\includegraphics[width=\textwidth]{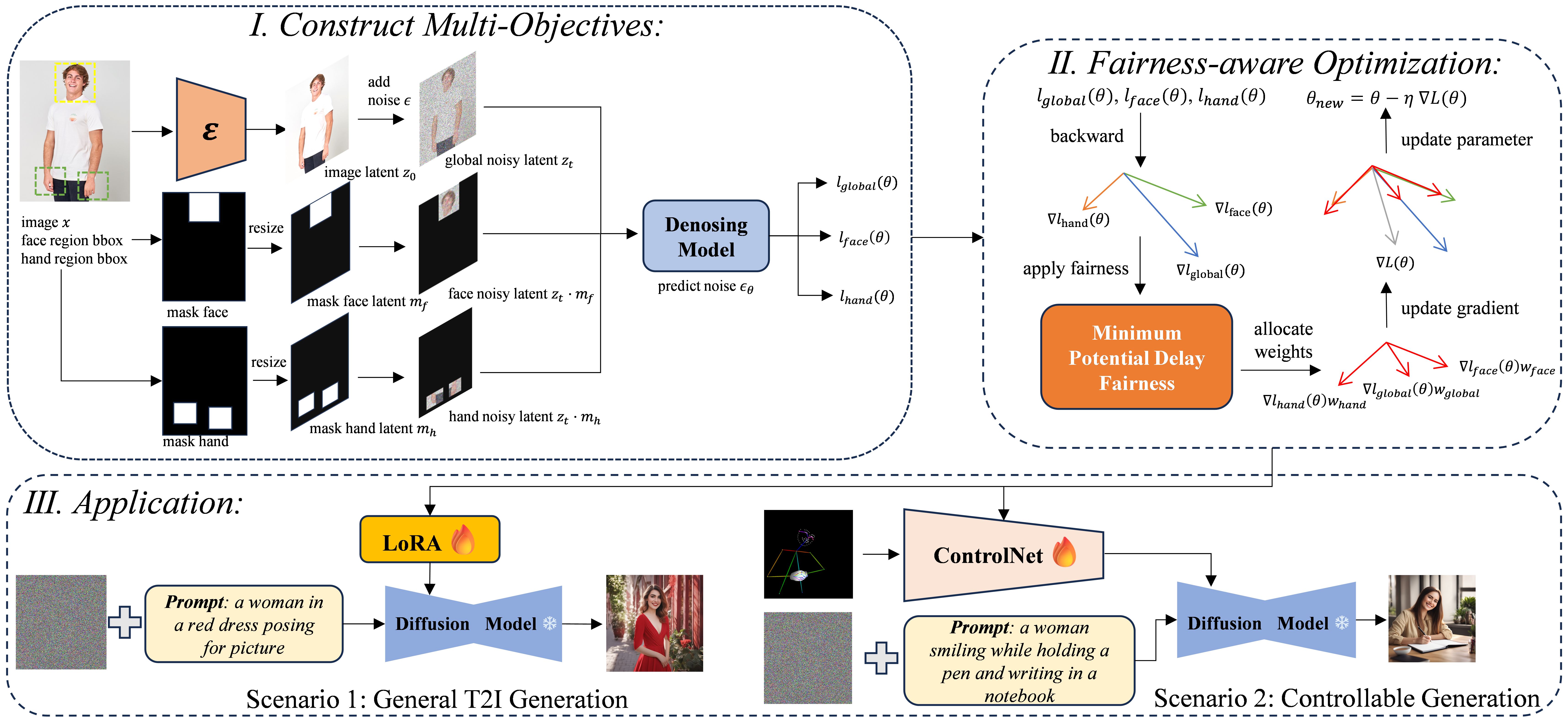}
\caption{\textbf{An overview of FairHuman framework.} The entire pipeline is divided into three stages: \textbf{1)} Construct multi objectives for human image generation; \textbf{2)} Employ fairness-aware optimization strategy for parameter updates; \textbf{3)} Apply the fine-tuned module into the backbone diffusion model. Here, we take LoRA\cite{hu2021lora} and ControlNet\cite{zhang2023adding} as examples of different generation scenarios. }
\label{fig:2}
\end{figure*}
\subsection{Problem Statement}\label{sec:problem formulation}
\noindent
The objective function of diffusion models is founded on the mean square error associated with the denoising task. It enhances the model's performance by minimizing the discrepancy between the predicted noise and the actual noise. In the case of Latent Diffusion Models (LDM)\cite{rombach2022high}, this denoising process is executed within the latent space, facilitated by a VAE encoder \cite{kingma2013auto}. This approach allows for more efficient and effective denoising by operating in a compressed representation of the data. Specifically, its optimization objective can be formulated as: 
\begin{equation}
\min _{\theta} E_{z_{0}, t, c,\epsilon}\left[\left\|\epsilon-\hat{\epsilon}_{\theta}\left(z_{t}, t,c\right)\right\|^{2}\right],\label{eq:mse_ori}
\end{equation}
where $z_t$ is the noisy latent at timestep $t$ obtained from 
$z_0$ through the diffusion process by adding Gaussian noise $\epsilon\sim N(0,1)$; 
the initial representation \( z_0 \) is defined as \( z_0 = \varepsilon(x_0) \), with \( \varepsilon \) denoting the encoder and \( x_0 \) being the ground truth image;
$\hat{\epsilon}_{\theta}\left(z_{t}, t,c\right)$ is the noise predicted by the diffusion model; $c$ represents the embedding of input conditions (\eg, textual prompt or image); $\theta$ represents the parameters to be updated. 

By pre-training on large-scale datasets with this objective function, models such as SDXL\cite{podell2023sdxl} have shown the ability to generate high-quality human content. This is evident in the first row of Fig.\ref{fig:3}, where close-up images of hands or faces are generated accurately. However, when generating full-body human images within a scene (see the second row of Fig.\ref{fig:3}), the model demonstrates inferior performance in rendering local details (faces and hands). In other words, when face or hand regions occupy a larger proportion of the image, their generation results are acceptable. On the contrary, when they occupy a smaller proportion of the image, their corresponding generation quality would be worse. 

What accounts for the occurrence of this issue? As indicated by Eq.(\ref{eq:mse_ori}), LDM calculates the loss based on the representation \( z_t \) encoded from the entire image without distinguishing the optimization difficulty of different regions. For faces and hands with larger areas in close-up images, the default objective function can be easily optimized since these regions contribute more significantly to the loss. Conversely, for human-in-the-scene images, the representation \( z_t \) comprises not only the hands and faces but also other body details and the background scene. Therefore, face and hand regions with smaller areas contribute less to the loss, thus leading to insufficient attention and resulting in poorer generation quality. \textbf{In this way, different regions (\eg, face, hands, global regions) may compete during the generation process.}
\subsection{Multi-objectives}\label{sec:3.2}
As we discussed above, the competition between different regions may lead to insufficient attention to some local areas, such as hands or faces. To address this issue, the generative model requires a fair allocation of resources to both local and global regions, thus enhancing the overall generation performance. To this end, we first construct the following optimization objectives for both local and global generation tasks, respectively:
\begin{equation}\label{eq:objective-functions}
\begin{alignedat}{2}
& l_{\text{global}}(\theta) &&= \mathbb{E}_{z_0, t, c, \epsilon} \left[ \left\| \epsilon - \hat{\epsilon}_{\theta}(z_t, t, c) \right\|^2 \right]; \vspace{0.3em} \\
&\  l_{\text{face}}(\theta) &&= \mathbb{E}_{z_0, t, c, \epsilon} \left[ \left\| m_f \odot (\epsilon - \hat{\epsilon}_{\theta}(z_t, t, c)) \right\|^2 \right]; \vspace{0.3em} \\
&\  l_{\text{hand}}(\theta) &&= \mathbb{E}_{z_0, t, c, \epsilon} \left[ \left\| m_h \odot (\epsilon - \hat{\epsilon}_{\theta}(z_t, t, c)) \right\|^2 \right],
\end{alignedat}
\end{equation}
where $\theta$ represents the learnable parameters; $l_{global}(\theta)$ is the global loss for entire image generation; $l_{face}(\theta)$ and $l_{hand}(\theta)$ are local losses focusing on faces and hands; $m_f$ and $m_h$ are the masks used to select the face and hand regions respectively. Here, the masks $m_f$ and $m_h$ are obtained by reshaping the mask image to the latent shape, following the same way as the inpainting mask $m$ denoted in Eq.(4) of \cite{lu2024handrefiner}. \textbf{For brevity, we also denote $l_{\text{global}}(\theta)$, $l_{\text{face}}(\theta)$, $l_{\text{hand}}(\theta)$ as $l_{1}(\theta)$, $l_{2}(\theta)$, $l_{3}(\theta)$ in the subsequent sections.} 

Based on this, we aim to fine-tune a diffusion model to simultaneously minimize $l_{1}(\theta)$, $l_{2}(\theta)$, and $l_{3}(\theta)$. Meanwhile, the fine-tuning approach should enhance the generation quality of hands and faces while maintaining its global performance. In this way, the fine-tuning task could be formulated as the following multi-objective optimization problem:
\begin{equation}\label{eq:mo_loss}
\min_{\theta} \mathcal{L}(\theta) = [l_{1}(\theta), l_{2}(\theta), l_{3}(\theta)]^\top.
\end{equation}
Notably, our primary goal of fine-tuning comprises two key aspects:  
\begin{itemize}
    \item \textbf{Regional Fairness:} We should ensure that the solution $\theta^*$ to Eq.(\ref{eq:mo_loss}) yields a fairly substantial improvement across all three objectives, especially for $l_{2}(\theta)$ and $l_{3}(\theta)$. As long as there is at least one insignificant improvement among the three objectives, it implies that our fine-tuning is ineffective. This aligns with our purpose of fair enhancement on the quality w.r.t face, hand, and global regions.
    \item \textbf{Pareto Optimality:} The solution $\theta^*$ to Eq.(\ref{eq:mo_loss}) should be Pareto optimal\cite{ngatchou2005pareto,sener2018multi}. Here we say a solution $\theta^*$ is Pareto optimal if there exists no other solution $\theta'$ satisfying $l_i(\theta') \leq l_i(\theta^*)$ for all $i \in \{1, 2, 3\}$. In other words, updating $\theta^*$ in any direction will disrupt at least one of the three objectives. Thus Pareto optimality here implies a trade-off between face, hand, and global quality.
\end{itemize}

\subsection{Fairness-aware Optimization}

We aim to ensure the post fine-tuning values of \(l_{1}(\theta)\), \(l_{2}(\theta)\), \(l_{3}(\theta)\) are all fairly minimized to a sufficient degree. To this end, a straightforward approach is to enable each \(l_{i}(\theta)\) could be maximally decreased at each training step of $\theta$. Formally, we seek a stepwise updating direction $d$ that maximizes the contributions of \(d\) to all of \(l_{1}(\theta)\), \(l_{2}(\theta)\) and \(l_{3}(\theta)\). From another perspective, this is equivalent to minimizing the negative effect of \(d\) to all these objectives.

Here, we employ the concept of \textbf{Potential Delay} \cite{massoulie1999bandwidth, mo2000fair} to quantitatively describe the negative effect. The term ``Potential Delay" originally refers to the expected time required for a user to complete a specific task in the context of resource allocation. For example, if an internet user transmits a unit-sized file at transmission rate $v$, then the potential delay is \({1}/{v}\). In this paper, the common updating direction \(d\) can be considered a shared resource among all objectives. The ``transmission rate" of this resource to each \(l_{i}(\theta)\) is represented by the projection magnitude of \(d\) onto the  steepest descent direction \(\nabla l_{i}(\theta)\), denoted as \(\|\text{proj}_{\nabla l_{i}(\theta)} (d)\|\). Consequently, the corresponding potential delay is \(1/\|\text{proj}_{\nabla l_{i}(\theta)} (d)\|\). Normally, projection \(\|\text{proj}_{\nabla l_{i}(\theta)} (d)\|\) can serve as a metric for the contribution of \(d\) to the optimization process of \(l_{i}(\theta)\) \cite{yu2020gradient}. In this way, the potential delay \(1/\|\text{proj}_{\nabla l_{i}(\theta)} (d)\|\) is inversely proportional to the contribution from \(d\) and thus directly proportional to the negative effect. In this way, we can determine the optimal direction \(d\) by resolving the subsequent minimization problem:
\begin{equation}\label{eq:min-proj}
\min_d \mathcal{F}(d)=\sum_{i=1}^{3} \frac{1}{\|\text{proj}_{\nabla l_{i}(\theta)} (d)\|}\ ,\\
\end{equation}
By optimizing Eq.(\ref{eq:min-proj}), the potential delay of \(l_{1}(\theta)\), \(l_{2}(\theta)\), \(l_{3}(\theta)\) can be minimized simultaneously, thus it is termed as \textbf{Minimal Potential Delay (MPD) Fairness} \cite{massoulie1999bandwidth, mo2000fair}. 

Prior to solving this optimization, we assume that \(d\) lies within a ball with radius $r$ centered at the origin. Based on this, the optimization problem w.r.t MPD fairness can be transformed into the following form:
\begin{equation}\label{eq:min-dot}
\min_d\mathcal{F}'(d)=\sum_{i=1}^{3} \frac{1}{\nabla l_{i}(\theta)^\top d}\ ,\\
\end{equation}
since that $\mathcal{F}'(d)$ is proportional to an upper bound of $\mathcal{F}(d)$ in Eq.(\ref{eq:min-proj}), and the proof is provided in the Sec.6.1 of the supplementary material. Here, $\mathcal{F}'(d)$ can be regarded as a special case of $\alpha$-fairness\cite{mo2000fair}. If we set $\alpha=2$ for Eq.(2) in \cite{ban2024fair}, it will be equivalent to Eq.(\ref{eq:min-dot}) in this paper. Thus, we can adopt a similar procedure to solve Eq.(\ref{eq:min-dot}) following \cite{ban2024fair}.

Specifically, we first apply Lagrange multiplier method on Eq.(\ref{eq:min-dot}) to obtain the following equation:
\begin{equation}\label{eq:6}
    d=\sum_{i=1}^{3} \frac{\nabla l_{i}(\theta)}{\left(\nabla l_{i}(\theta)^{\top} d\right)^{2}} ,
\end{equation}
The derivation process is described in Sec.6.2 of the supplementary material. To further simplify the problem, we follow \cite{ban2024fair} and express \(d\) as a linear combination:
\begin{equation}\label{eq:dw}
    d = \sum_{i=1}^{3} w_i \nabla l_{i}(\theta) ,
\end{equation}
where \(w_i\) represents the weight of the \(i\)-th objective. Given that each gradient $\nabla l_{i}(\theta)$ is readily available, the problem of determining \(d\) is reformulated as determining every $w_i$. By substituting Eq.(\ref{eq:dw}) into Eq.(\ref{eq:6}), we can get that:
\begin{equation}\label{eq:7}
   G(\theta)^{\top}G(\theta) W=W^{-1/2} ,
\end{equation}
where $G(\theta)=[\nabla{l_1}(\theta),\nabla{l_2}(\theta), \nabla{l_3}(\theta)]^\top$ and $W=[w_{1},w_{2},w_{3}]^\top$. According to the relation in Eq.(\ref{eq:7}), we can further construct the following convex optimization problem:
\begin{equation}\label{eq:nls}
   \min _W\left\|G(\theta)^{\top} G(\theta) W-W^{-1/2}\right\|^2 ,
\end{equation}
By applying the standard least square method, we can obtain the following close-form solution of Eq.(\ref{eq:nls}):
\begin{equation}\label{eq:solution_W}
   W = \left( G(\theta)^\top G(\theta) \right)^{-\frac{2}{3}}\mathbf{1} ,
\end{equation}
where $\mathbf{1}$ represents an all-ones vector. With these weights, we can finally obtain the value of $d$ by using Eq.(\ref{eq:dw}) and then conduct an updating step as: 
\begin{equation}\label{eq:parameter update}
\begin{array}{l}
\theta_{new}=\theta-\eta d,
\end{array}
\end{equation}
where $\theta_{new}$ is the updated parameter and $\eta$ is the learning rate. By adopting this approach, each training step is guided by MPD fairness, thereby ensuring the efficacy of the fine-tuning results across all objectives. Meanwhile, Thm.(7.3) in \cite{ban2024fair} also indicates that such a training process can converge to a Pareto optimal solution. And our constructed objective functions in Eq.(\ref{eq:objective-functions}) decouples regional and global optimization, satisfying MPD’s prerequisite assumption of linearly independent task gradients.

\subsection{Application}\label{sec:application}
Considering the high cost of full fine-tuning for diffusion models, we opt to deploy our method in LoRA\cite{hu2021lora} and ControlNet\cite{zhang2023adding} while freezing the parameters of the backbone model. They represent two types of generation patterns, namely, general T2I generation and controllable generation with control signals. Specifically, LoRA primarily enhances the quality of human images from the perspective of visual concepts and pixel information, while ControlNet focuses on more refined and controllable generation over the structural aspects. Both have significant influence and play important roles in the field of human image generation. We construct a high-quality dataset for their fine-tuning. Apart from the images and text descriptions, it also includes the positional information for creating mask-based objective functions and the pose-depth information for controllable generation. Their new outputs can be formulated as:
\begin{equation}\label{eq:output}
\hat{y}_{new}=N_{base}\left(z ; \theta_{base}\right)+\beta N_{new}\left(c, z ; \theta_{update}\right) ,
\end{equation}
where $z$ is the input latent and $N_{base}$ is the backbone network with the frozen parameters $\theta_{base}$. $ N_{new}$ represents the fine-tuned module (LoRA or ControlNet here) with the updated parameters $\theta_{update}$. As for condition $c$, it contains textual and timestep embeddings for the LoRA module and has additional control signals for the ControlNet module. $\beta$ is the hyperparameter used for scaling.

%% file: sec/4_experiments.tex
\section{Experiments}
\label{sec:Experiments}
\subsection{Experimental Settings}\label{sec:experimental settings}
\textbf{Datasets.} We collect human-centric images from two public datasets called CosmicMan\cite{li2024cosmicman} and BodyHands\cite{narasimhaswamy2022whose}. To enable effective training, we implement several preprocessing methods, including filtering, captioning, and labeling. Ultimately, we construct a high-quality human image dataset that contains approximately 50k images for training. As for the validation, we randomly sample about 5k in-the-wild human images under different scenes with well-processed captions provided by \cite{zhu2024mole}. The details regarding the construction of our dataset will be elaborated in Sec.7 of the supplementary material.\\
\textbf{Baseline Methods.} We compare our approach with two categories of baseline methods. \textbf{1)} General T2I methods without extra conditions: SDXL-base\cite{podell2023sdxl} and MoLE\cite{zhu2024mole}. Notably, MoLE is a recent work based on the collaboration of multiple LoRA modules. \textbf{2)} Controllable methods with extra conditions: ControlNet-SDXL\cite{zhang2023adding} and ControlNet Union\cite{li2024controlnet++}. Here, we compare performance under the same given conditions (depth and pose) using different training strategies.
\begin{table*}[t]
\Huge
\vspace{5pt}
\centering
\resizebox{1.0\linewidth}{!}{
\begin{tabular}{llccccccc}
\toprule
& & \multicolumn{2}{c}{Image Quality} & \multicolumn{1}{c}{Text-Image Alignment} & \multicolumn{3}{c}{Regional Quality} & \multicolumn{1}{c}{Pose Alignment} \\
\cmidrule(l){3-4} \cmidrule(l){5-5} \cmidrule(l){6-8} \cmidrule(l){9-9}
\multirow{-2}{*}{Application Scenario} & \multirow{-2}{*}{Model} & HPS(\%) $\uparrow$ & IR(\%) $\uparrow$ & CLIP $\uparrow$ & FID $\downarrow$ & Hand Confi. $\uparrow$ & Face Confi. $\uparrow$ & Mean Diff $\downarrow$ \\
\toprule
\multirow{3}{*}{\textbf{General T2I Methods}} 
& SD-XL Base \cite{podell2023sdxl} & 30.91 & 140.85 & 26.84 & 52.69 & 93.18 & 83.67 & / \\
& MoLE \cite{zhu2024mole} & 31.87 & 152.36 & 27.17 & 48.97 & 93.32 & 84.48 & /  \\
& Ours  & \textbf{32.73} & \textbf{154.60} & \textbf{27.20} & \textbf{46.11} & \textbf{94.03} & \textbf{85.31} & / \\
\toprule
\multirow{3}{*}{\textbf{Controllable Methods}}
& ControlNet-SDXL \cite{zhang2023adding} & 32.20 & 152.49 & 27.10 & 46.18 & 93.50 & 85.13 & 3.45 \\
& ControlNet-Union \cite{li2024controlnet++} & \textbf{32.80} & \textbf{155.22} & 27.21 & 46.43 & 94.28 & 85.36 & 2.67 \\
& ControlNet-Ours  & 32.72 & 154.74 & \textbf{27.24} & \textbf{45.46} & \textbf{94.48} & \textbf{85.61} & \textbf{2.35} \\
\bottomrule
\end{tabular}
}
\caption{\textbf{The evaluation results for ours and other existing methods.} We conduct comparisons under two different image generation scenarios separately. As for the general T2I methods, we compare our LoRA-based method with a widely adopted diffusion model SDXL\cite{podell2023sdxl} and a recent work MoLE\cite{zhu2024mole}. For controllable methods, we compare our ControlNet-based method with the default ControlNet-SDXL\cite{zhang2023adding} and the latest ControlNet-Union\cite{li2024controlnet++}. Notably, the mean difference of pose condition is employed here to evaluate the effectiveness of controllable generation. The best results are indicated in bold.}
\label{tab1}
\end{table*}
\\
\noindent
\textbf{Metrics.} We adopt evaluation metrics based on the optimization objectives outlined in the paper, focusing on two dimensions. \textbf{1)} Overall image quality: Human Preference Score (HPS)\cite{wu2023human} and ImageReward (IR)\cite{xu2023imagereward}. They are built on the CLIP\cite{radford2021learning} and BLIP\cite{li2022blip} models respectively to reflect human visual preferences from multiple aspects including image alignment, aesthetic, and plausibility. Moreover, CLIP score is applied to evaluate text-image alignment, and pose alignment is used for assessing the effectiveness of controllable generation. \textbf{2)} Regional generation quality: FID\cite{heusel2017gans} and detection confidence of mediapipe\cite{lugaresi2019mediapipe} are utilized to evaluate the quality of hands and faces in terms of realism and plausibility. The implementation details about our experiments are provided in Sec.8 of the supplementary material.
\subsection{Comparison with Existing Methods}
\textbf{Quantitative Analysis.} The evaluation results compared with existing methods are shown in Tab.\ref{tab1}. We perform evaluations under two classic scenarios. Specifically,  for general T2I methods, including SDXL-Base\cite{podell2023sdxl}, MoLE\cite{zhu2024mole}, and our LoRA-based method, we randomly sample 5k prompts about human-in-the-scene images from the validation dataset and feed them to each model for image generation. As for the setting of hyperparameters, we keep the default CFG scale of 7.5, set the inference step to 30, and fix the seed at 0 for all methods. In particular, for our LoRA-based method, we implement three different weights (0.3, 0.4, and 0.5) in the experiments and use their average metric values as the final result. To more comprehensively evaluate the quality of generated images, we utilize the metrics in terms of overall image quality, text-image alignment, and regional quality. The results demonstrate that our method outperforms existing approaches in all these three aspects, especially in the generation quality of local hand and face details. Since our LoRA-based method and MoLE are deployed on SDXL sharing the same backbone model, we also conduct experiments on their memory usage and inference speed. Results suggest that our method can achieve better performance while requiring less memory size and inference time. Related details are demonstrated in Sec.9.1 of the supplementary material.

\begin{figure}[ht]
  \centering
   \includegraphics[width=1.0\linewidth]{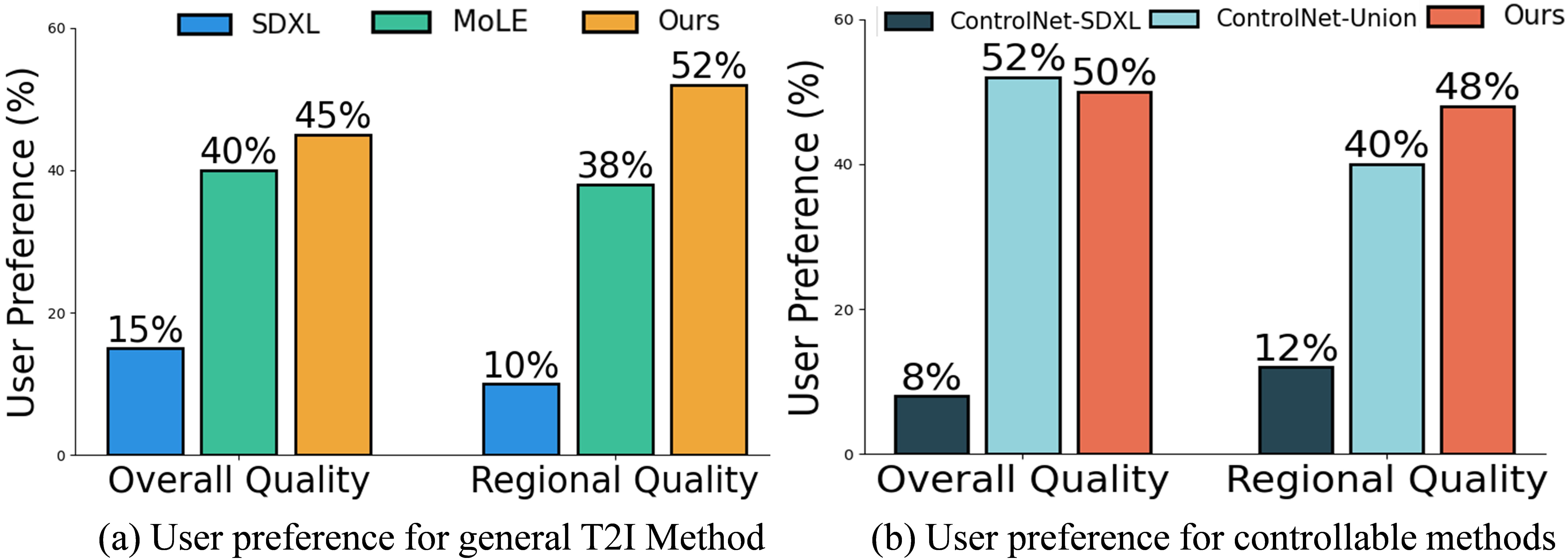}
   \caption{\textbf{User study results in terms of overall quality and regional quality.} We present the ratio of users' preference for each method.}
   \label{fig:5}
\end{figure}

Furthermore, to verify the effectiveness of our method across different task scenarios, we employ it in ControlNet\cite{zhang2023adding} for the controllable generation. As shown in the lower part of Tab.\ref{tab1}, we compare with existing controllable methods, including the default ControlNet-SDXL and the latest variant ControlNet-Union\cite{li2024controlnet++}.
The evaluation is conducted using the same metrics as described above. Similar to our LoRA-based method, we set three different control scales (0.4, 0.5, and 0.6) here for testing and choose their average metric values. Besides, considering the characteristics of controllable generation, we also evaluate the model's performance based on pose alignment, \ie, the consistency between the input conditions and the actual generated content. Here, we use the mean difference between the input and extracted conditions as the metric. The corresponding results are shown in the last column of Tab.\ref{tab1}. In comparison, our method still demonstrates superior performance in the controllable generation scenario. Although it is slightly inferior to ControlNet-Union in terms of the HPS and IR metrics, the primary reason is that ControlNet-Union incorporates aesthetic and preference feedback learning during training, whereas our method focuses more on the realism and plausibility of the images. Meanwhile, considering our framework’s property of extensibility, it is reasonable to introduce human preference as a new sub-task in future work. Furthermore, as demonstrated in Tab.\ref{tab1} and Fig.\ref{fig:5}(b), our approach achieves comparable overall quality with only marginal differences while significantly outperforming in regional details.
\begin{table*}[t]
\vspace{5pt}
\centering
\resizebox{1.0\linewidth}{!}{
\begin{tabular}{llcccccc}
\toprule
& & \multicolumn{2}{c}{Image Quality} & \multicolumn{3}{c}{Regional Quality} & \multicolumn{1}{c}{Pose Alignment} \\
\cmidrule(l){3-4}  \cmidrule(l){5-7} \cmidrule(l){8-8}
\multirow{-2}{*}{Model} & \multirow{-2}{*}{Ablation Settings} & HPS(\%) $\uparrow$ & IR(\%) $\uparrow$ & FID $\downarrow$ & Hand Confi. $\uparrow$ & Face Confi. $\uparrow$ & Mean Diff $\downarrow$ \\
\toprule
\multirow{3}{*}{\textbf{LoRA}} 
& $l_{1}(\theta)$ & 32.40 & 151.08 & 49.48 & 92.81 & 83.84 & / \\
& $l_{1}(\theta)+l_{2}(\theta)+l_{3}(\theta)$ (w/o fairness)& 32.60 & 153.23 & 47.22 & 93.70 & 84.82 & /  \\
& $l_{2}(\theta)+l_{2}(\theta)+l_{3}(\theta)$ (w fairness) & \textbf{32.73} & \textbf{154.60}  & \textbf{46.11} & \textbf{94.03} & \textbf{85.31} & / \\
\toprule
\multirow{3}{*}{\textbf{ControlNet}}
&$l_{1}(\theta)$ & 32.20 & 152.49 & 46.18 & 93.50 & 85.13 & 3.45 \\
& $l_{1}(\theta)+l_{2}(\theta)+l_{3}(\theta)$ (w/o fairness)& 32.56 & 154.30 & 46.79 & 93.99 & 85.24 & 2.78 \\
&$l_{1}(\theta)+l_{2}(\theta)+l_{3}(\theta)$ (w fairness) & \textbf{32.72} & \textbf{154.74} & \textbf{45.36} & \textbf{94.48} & \textbf{85.61} & \textbf{2.35} \\
\bottomrule
\end{tabular}
}
\caption{\textbf{Ablation results deployed on LoRA and ControlNet.} We primarily explore the effect of two aspects: additional objective functions for regional generation and the learning strategy. $l_{1}(\theta)$, $l_{2}(\theta)$, and $l_{3}(\theta)$ are defined in Eq.\ref{eq:objective-functions}. The best results are indicated in bold.}
\label{tab:2}
\end{table*}
\noindent
\\
\textbf{Qualitative Analysis.} Fig.\ref{fig:1} shows qualitative results. Compared with baseline methods, our method can generate human images with higher quality in both global and local aspects. Meanwhile, it achieves better controllability and plausibility in terms of controllable generation. We also conduct a user study to provide a more intuitive evaluation. Volunteers are invited to select images that better align with their visual preferences based on overall and regional quality. The result is shown in Fig.\ref{fig:5}. It indicates that our method is more preferred by users in both two aspects. More details about the user study will be demonstrated in Sec.9.2 of the supplementary material.
\begin{table}[t]
\Huge
\vspace{5pt}
\centering
\resizebox{1.0\linewidth}{!}{
\begin{tabular}{lcccccc}
\toprule
& \multicolumn{2}{c}{Image Quality} & \multicolumn{1}{c}{T2I Alignment} & \multicolumn{3}{c}{Regional Quality} \\
\cmidrule(l){2-3} \cmidrule(l){4-4} \cmidrule(l){5-7}
\multirow{-2}{*}{LoRA Rank} & HPS(\%) $\uparrow$ & IR(\%) $\uparrow$ & CLIP$\uparrow$ &FID $\downarrow$ & Hand Confi. $\uparrow$ & Face Confi. $\uparrow$ \\
\toprule
r=64 & 32.50 & 148.52& 27.11& 47.87 & 93.25 & 84.69 \\
r=128 &\textbf{32.73} & \textbf{154.8} &27.15 & 46.79 & 93.79 & 85.11 \\
r=256  & \textbf{32.73} & 154.60 & \textbf{27.20}&\textbf{46.11} & \textbf{94.03} & \textbf{85.31} \\
\bottomrule
\end{tabular}
}
\caption{\textbf{Comparison results for the choice of LoRA rank.} The best results are indicated in bold.} 
\label{tab:ablation rank}
\end{table}
\begin{figure}[ht]
 \centering
  \includegraphics[width=1.0\linewidth]{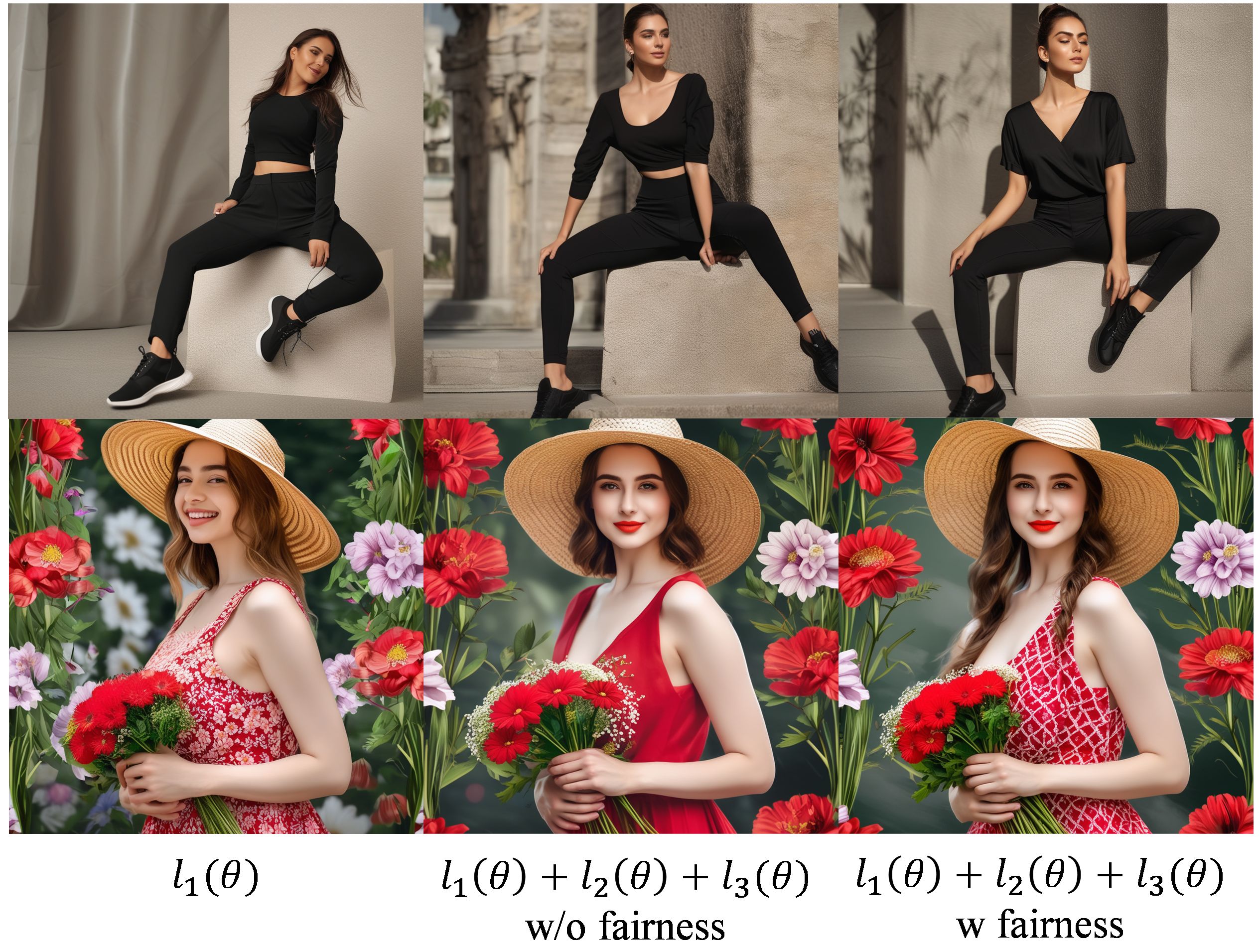}
  \caption{\textbf{Examples of the images generated by different settings with the same prompt.} First row: generated by LoRA. Second row: generated by ControlNet. Zoom in for a better view of details.}
  \label{fig:6}
\end{figure}
\subsection{Ablation Studies}
We present the key ablation studies based on the construction of our objective functions and the strategy for optimization. Here, we analyze their effects according to the overall and local quality of the generated images, deploying on LoRA and ControlNet, respectively. We randomly select 1k human-related prompts and follow the same hyperparameter setting as mentioned above for testing. Their results are shown in Tab.\ref{tab:2}. In addition, some visualization results are provided in Fig.\ref{fig:6}.\\
\textbf{Effect of Regional Optimization Objective.} To explore how regional loss functions including $l_{face}(\theta)$ and $l_{hand}(\theta)$ in Eq.(\ref{eq:objective-functions}) influence generation performance, we fine-tuned the corresponding LoRA and ControlNet by adopting the default loss function from Eq.(\ref{eq:mse_ori}) and the joint objective functions in Eq.(\ref{eq:objective-functions}), respectively. For the latter, we employ a fixed-weight linear scalarization method here, transforming the multi-objective optimization into a single-objective optimization. The first two rows for each model in Tab.\ref{tab:2} present the evaluation results of the two approaches, validating that the inclusion of regional loss functions helps the model improve its generation capabilities for details such as hands and faces, thereby enhancing the overall quality of the images.\\
\textbf{Effect of Fairness Principle.} In our work, we utilize the minimum potential delay (MPD) fairness to achieve better optimization based on the joint of objective functions for global and regional generation. Here, we compare two gradient update strategies: one without considering fairness (\ie, a fixed weight allocation based on linear scalarization method) and one considering fairness (\ie., a dynamical weight allocation based on MPD fairness). As shown in the last two rows for each model in Tab.\ref{tab:2}, the optimization strategy based on the MPD fairness can better focus on the more challenging objectives of generating faces and hands without compromising the overall quality. It achieves a more balanced performance among different objectives. 
\subsection{More Analysis}
Considering that the rank setting significantly impacts the LoRA performance, we additionally conduct a hyperparameter study to figure out the optimal rank configuration. The results are presented in Tab.\ref{tab:ablation rank}, which indicates that setting the rank value to 256 maximizes the model’s ability to generate regional details without excessively jeopardizing the overall image quality. Besides, we also conduct comparison studies on some other multi-objective learning strategies, \eg, Dynamic Weight Average (DWA)\cite{liu2019end}, Uncertainty Weighting (UW)\cite{kendall2018multi}, and Random Loss Weighting (RLW)\cite{lin2021reasonable}. The results suggest that applying the MPD fairness principle to the optimization process can achieve a more balanced and superior performance in terms of the human image generation task. Relevant results and details are presented in Sec.9.3 of the supplementary material.
\subsection{Visualizations}
We put more visualization results compared with existing methods in Sec.10 of the supplementary material, including other close-source methods like HyperHuman\cite{liu2023hyperhuman} and HanDiffuser\cite{narasimhaswamy2024handiffuser}. Our method still demonstrates comparable or superior performance.

%% file: sec/5_conclusion.tex
\section{Conclusion}
In this work, we propose FairHuman to enhance the quality of human images generated by diffusion models. To address the distortion issue in local details like faces and hands, we specifically construct corresponding objective functions for regional generation. Then, we combine the default global objective function to achieve joint training. The key of our method is regarding the improvement of human image generation as a multi-objective optimization task. By utilizing the minimum potential delay principle and the gradient manipulation process, our method enables more balanced and effective performance. Extensive experiments demonstrate the superiority of our method under different scenarios of human image generation. For future work, we plan to further design and optimize corresponding objective functions for more attributes related to humans, such as feet and eyes.

%% file: sec/X_suppl.tex
\clearpage
\setcounter{page}{1}
\maketitlesupplementary
\section{Proofs}
\label{sec:proofs}

\subsection{Derivation of Eq.(\ref{eq:min-dot})}
\label{sec:ub}
Assuming there exists a constant \( M \) such that \( \|\nabla l_i(\theta)\| \leq M \), we have:
\[
\begin{aligned}
\mathcal{F}(d) &=\sum_{i=1}^{3} \frac{1}{\|\text{proj}_{\nabla l_{i}(\theta)} (d)\|}\ \\
&=\sum_{i=1}^{3} \frac{1}{\nabla l_{i}(\theta)^\top d / \|d\|}\ \\
&=\sum_{i=1}^{3} \frac{\|d\|}{\nabla l_{i}(\theta)^\top d }\ \\
&\leq \sum_{i=1}^3 \frac{M}{\nabla l_{i}(\theta)^\top d} \\
&= M \sum_{i=1}^3 \frac{1}{\nabla l_{i}(\theta)^\top d}
\end{aligned}
\]
Since \( \mathcal{F}'(d) = \sum_{i=1}^3 \frac{1}{\nabla l_{i}(\theta)^\top d} \), we can get that:

\[
\mathcal{F}(d) \leq M \mathcal{F}'(d)
\]
Therefore, \( \mathcal{F}'(d) \) is proportional to an upper bound of \( \mathcal{F}(d) \), with the proportionality constant being \( M \).
\subsection{Derivation of Eq.(\ref{eq:6})}\label{sec:extremum proof}
Now we have the following optimization objective:
\begin{equation}
\min_d\mathcal{F}'(d)=\sum_{i=1}^{3} \frac{1}{\nabla l_{i}(\theta)^\top d}\ ,\\
\text { s.t. }
\|d\|^{2} \leq r^{2}
\end{equation}
where $\nabla l_{i}(\theta)$ represents the gradient of each objective function $l_{i}(\theta)$ and $d$ is the update direction for all objectives. $r$ is the radius of the boundary space. Notably, the objective function above is non-decreasing for any feasible $d$. Therefore, for any $d$ in the boundary space, there must exist a point in the same direction but on the boundary. For the maximization of utility, we can conclude that the optimal $d^*$ must lie on the boundary, \ie, $\|d^*\|=r$. To solve the extremum problem above, we apply Lagrange multiplier method in the following steps:
\begin{itemize}
    \item \textbf{Construct Lagrange function.} By introducing the Lagrange multiplier $\lambda$, we can obtain the Lagrangian function:
    \begin{equation}
\begin{array}{ll}
\mathcal{L}(d, \lambda)=\sum_{i=1}^{3}\frac{1}{\nabla l_{i}(\theta)^{\top} d}+\lambda\left(\|d\|^{2}-r^{2}\right)
\end{array}
\end{equation}
    \item \textbf{Take the derivative and set it equal to zero.} Take the partial derivatives of $\lambda$ and $d$, set them equal to zero:
    \begin{equation}\label{eq:derivatives}
\begin{array}{ll}
\frac{\partial \mathcal{L}}{\partial d}=-\sum_{i=1}^{3} \frac{\nabla l_{i}(\theta)}{\left(\nabla l_{i}(\theta)^{\top} d\right)^{2}}+2 \lambda d=0 \\
\frac{\partial \mathcal{L}}{\partial \lambda}=\|d\|^{2}-r^{2}=0
\end{array}
\end{equation}
    \item \textbf{Solve the equation.} From the first formula in Eq.(\ref{eq:derivatives}), we can obtain:
    \begin{equation}
\begin{array}{ll}
d=\frac{1}{2\lambda}\sum_{i=1}^{3} \frac{\nabla l_{i}(\theta)}{\left(\nabla l_{i}(\theta)^{\top} d\right)^{2}}
\end{array}
\end{equation}
\end{itemize}
Here, $\lambda$ is set to $\frac{1}{2}$ for simplicity following \cite{navon2022multi}, which has no impact on the final results. It indicates that the update direction $d$ can be regarded as a weighted sum of gradient $\nabla l_{i}(\theta)$. The weight is allocated depending on the dot product of the gradient and the current update direction. Through this method, we can find the optimal update direction within the boundary space.

\section{Dataset Details }
\label{sec:dataset details}
\subsection{Construction of Training Set}\label{sec:training set}
About the two public datasets we use, CosmicMan is specifically designed for generating highly realistic and photorealistic human images, containing 6 million high-resolution real-world human images and detailed descriptions of 115 million diverse attributes, which helps the model learn a wide range of human details and scene information. BodyHands focuses on human poses, particularly the torso and hands, providing additional information on the structure of the human body. To enable effective training, we utilize state-of-the-art detectors\cite{yang2023effective, pavlakos2024reconstructing} to filter low-resolution images (smaller than 1024$\times$1024) or unrelated images (containing little information about humans). Furthermore, we apply visual language models (VLMs)\cite{gao2024mini,hu2024minicpm} to achieve accurate and concise image caption generation. For each sample annotation, in addition to the text description, we add extra positional information based on the detection results for our target optimization regions, namely the faces and hands, while also providing pose and depth information for the controllable generation.
\subsection{Construction of Validation Set}\label{sec:validation set}
To provide a more comprehensive evaluation of the model's performance in human image generation, we construct our validation set based on the large-scale human-centric image dataset proposed by MoLE\cite{zhu2024mole}, which mainly includes the following characteristics:
\begin{itemize}
    \item \textbf{Brief and clear text prompts:} All text prompts are processed through VLMs and manual review to remove information unrelated to the image content, such as complex adjectives and clauses, retaining only the key information related to humans. For example: "A woman in dress standing in front of a building", "A man in a suit and tie standing with his hands on his hips". This ensures that the model does not generate content inconsistent with the target, thereby avoiding a negative impact on the final evaluation.
    \item \textbf{Diverse content and scenarios:} To cover human images in various scenarios and content as comprehensively as possible, we randomly sample 5k images and their corresponding text descriptions from the 23411 selected human-in-the-scene images for each batch testing. This includes different human races, genders, and scenarios.
    \item \textbf{High-quality ground truth images:} To ensure the quality of the image data, we filter the existing dataset based on resolution, clarity, and content relevance.
\end{itemize}

\section{Implementation Details }
\label{sec:implementation details}
We choose Stable Diffusion XL as the base model since it already possesses prior knowledge about human image generation through pre-training but has major limitations in generating local hand and face details. Regarding the generation of local masks for the faces and hands, we first generate mask images using the pre-annotated bounding box information from the dataset. Then, we apply the same pre-processing methods as used for the original image, including resizing, cropping, and random flipping. Our code is developed based on Diffusers. For the LoRA fine-tuning, we set the rank to 256 and train it for 50k steps with a learning rate at 1e-5. The Adam optimizer is deployed. Compared to the default settings, we mainly increase the rank size while reducing the learning rate, as we expect the model to learn detailed content better. For the ControlNet fine-tuning, we train it 60k steps with a learning rate set to 1e-5. It is worth mentioning that we apply a dropout probability of 0.3 to the input control conditions to enhance robustness and generalization. The overall image resolution is set as 1024 × 1024. Our resource-friendly training and evaluation processes can be implemented on a single 80G NVIDIA A800 GPU.

\section{More Evaluation Details }
\label{sec:more evaluation details}
\subsection{More Quantitative Comparisons}\label{sec:more quantitative comparisons}
In addition to comparing the image quality generated by the models, we also conduct experiments on memory usage and inference speed. Specifically, both our LoRA-based method and MoLE\cite{zhu2024mole} are deployed on SDXL\cite{podell2023sdxl} as the backbone model. Therefore, here we primarily present the extra memory consumption and inference time required per image (set inference steps to 30 and employ UniPCMultistepScheduler as the sampler). The results are demonstrated in Tab.\ref{tab:more quantitative comparisons}.
\begin{table}[ht]
\centering
\begin{tabular}{lll} 
\hline
\textbf{Methods} & \textbf{Extra Memory Usage} & \textbf{Inference Time}  \\ 
\hline
MoLE             & 2611.75MB                   & ×3                       \\
Ours             & 1413.12MB                   & ×1.4                     \\
\hline
\end{tabular}
\caption{Comparisons between MoLE and our method in terms of memory usage and inference speed.}
\label{tab:more quantitative comparisons}
\end{table}
In particular, MoLE uses two low-rank modules and a gate network to achieve adaptive generation. Our method, on the other hand, achieves joint optimization of multiple objectives through a single low-rank module, thereby reducing memory requirements by nearly half and also significantly reducing the time required for inference.
\subsection{User Study Details}\label{sec:user study}
The user study involves 50 participants to evaluate 100 pairs of images in total with corresponding annotations generated by different methods. Images with irrelevant content are pre-filtered and removed. In complying with the quantitative analysis, participants are asked to rate them according to the following two criteria, respectively:
\begin{itemize}
    \item \textbf{Overall quality:} assessing the general appearance, realism, and coherence of the entire image.
    \item \textbf{Regional quality:} evaluating specific regions of interest (faces and hands) for detail, plausibility, and naturalness.
\end{itemize}
We note that all the participants are unaware of which image corresponds to which method and rank the images based on their preferences. For the highest rank in each group, we record its score as 1 and the rest as 0. In addition, for tied rankings, we assign a score of 0.5 to each. Finally, we separately calculate the scores for each method based on the two aforementioned criteria and visualize them using the bar chart, which provides a more comprehensive and intuitive reflection of the image generation quality.
\subsection{More Ablation Studies}\label{sec:more ablation studies}
For our LoRA-based method, we conduct additional ablation studies on the choice of rank and multi-objective optimization strategy. The results are demonstrated in Tab.\ref{tab:ablation rank} and Tab.\ref{tab:ablation strategy}, respectively.\\
\textbf{LoRA Rank.} 
As for the choice of LoRA rank, we experiment with three settings: 64, 128, and 256. Generally, a larger rank means more trainable parameters are introduced, enhancing the model's adaptability to new data. However, this also increases computational and memory demands, potentially leading to over-fitting. Therefore, we aim to identify the most suitable parameter choice for human image generation through comparative experiments.\\
\textbf{Multi-Objective Optimization Strategy.} 
We further deploy several classic multi-task learning strategies during the training process and conduct related comparative experiments. Specifically, Linear
Scalarization (LS), Dynamic Weigth Average (DWA)\cite{liu2019end}, Uncertainty Weighting (UW)\cite{kendall2018multi}, Random Loss Weighting (RLW)\cite{lin2021reasonable}, Scale-Invariant (SI), Nash-MTL\cite{navon2022multi}, and  Minimum Potential Delay Fairness Grad (MPD-FairGrad) are implemented here. Notably, SI utilizes the proportional fairness principle. Results in Tab.\ref{tab:ablation strategy} demonstrate that utilizing the MPD fairness principle to the multi-objective optimization for human image generation can achieve a more balanced performance.
\begin{table}[ht]
\Huge
\vspace{5pt}
\centering
\resizebox{1.0\linewidth}{!}{
\begin{tabular}{lccccc}
\toprule
& \multicolumn{2}{c}{Image Quality} & \multicolumn{3}{c}{Regional Quality} \\
\cmidrule(l){2-3} \cmidrule(l){4-6}
\multirow{-2}{*}{Strategy} & HPS(\%) $\uparrow$ & IR(\%) $\uparrow$ & FID $\downarrow$ & Hand Confi. $\uparrow$ & Face Confi. $\uparrow$ \\
\toprule
LS & 32.60 & 153.23 & 47.22 & 93.70 & 84.82 \\
DWA & 32.53 & 152.58 & 46.34 & 93.40 & 84.86 \\
UW & 32.61 & 152.82 & 47.42 & 93.76 & 84.15 \\
RLW & 32.58 & 154.56 & 47.90 & 93.80 & 84.24 \\
SI & 32.52 & 154.58 & \textbf{46.01} & 93.99 & 85.31 \\
Nash-MTL & 32.63 & 153.11 & 47.50 & 93.68 & 84.33 \\
MPD-FairGrad & \textbf{32.73} & \textbf{154.60} & 46.11 & \textbf{94.03} & \textbf{85.34} \\
\bottomrule
\end{tabular}
}
\caption{Ablation results for the choice of multi-objective optimization strategy.} 
\label{tab:ablation strategy}
\end{table}

\section{More Visualizations}
\label{sec:more visualizations}
\begin{figure*}[htb]
\centering
\includegraphics[width=\textwidth]{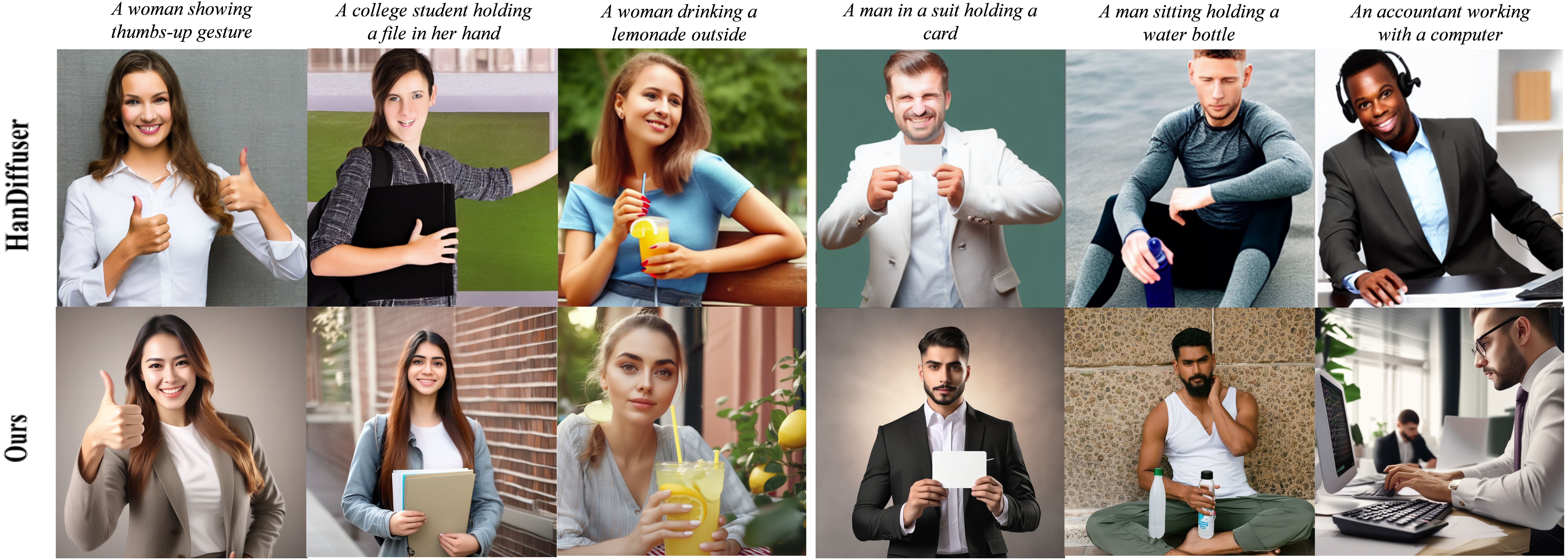}
\caption{Comparison with HanDiffuser.}
\label{fig:s1}
\end{figure*}

\begin{figure*}[htb]
\centering
\includegraphics[width=\textwidth]{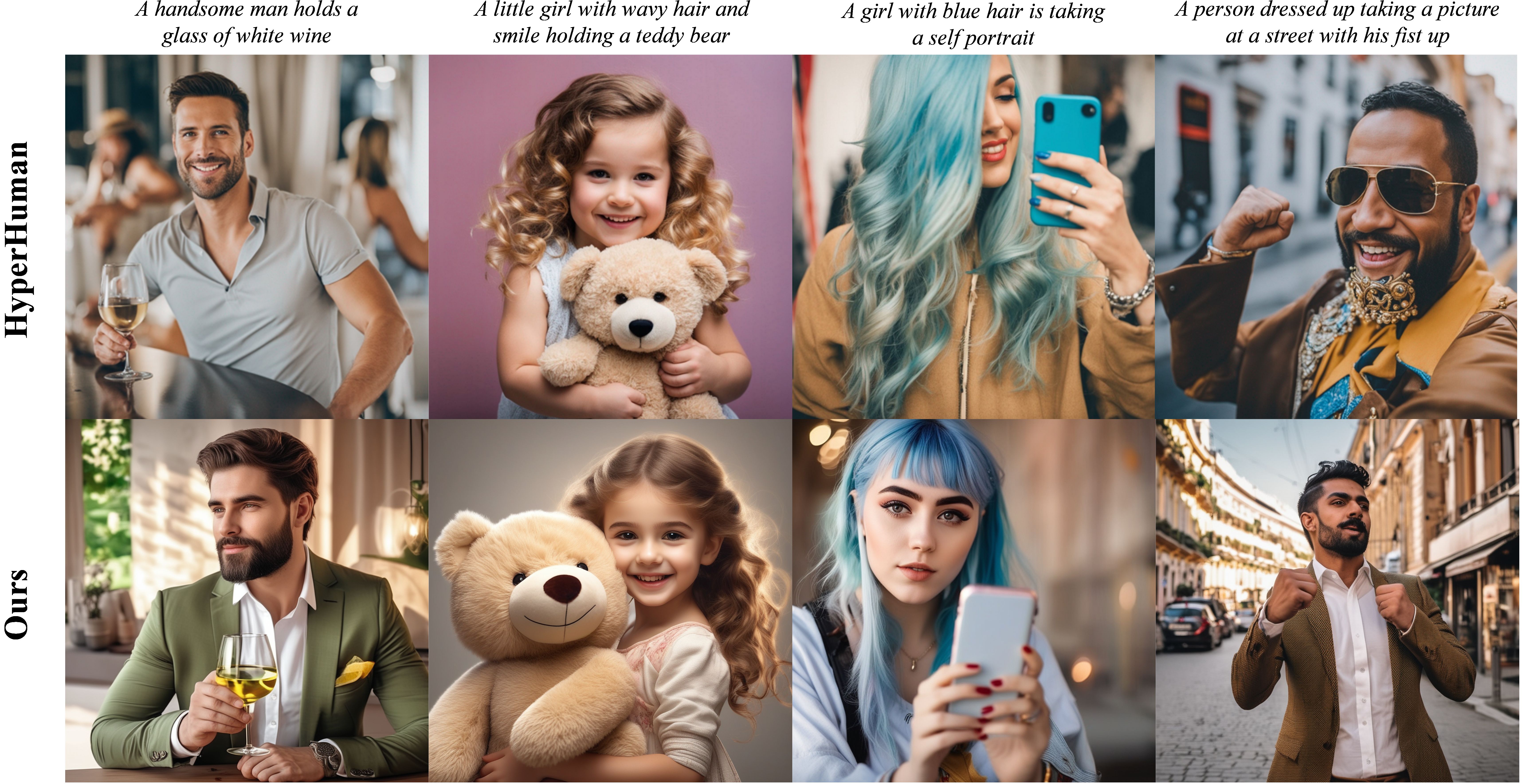}
\caption{Comparison with HyperHuman.}
\label{fig:s2}
\end{figure*}

\begin{figure*}[htb]
\centering
\includegraphics[width=1\textwidth]{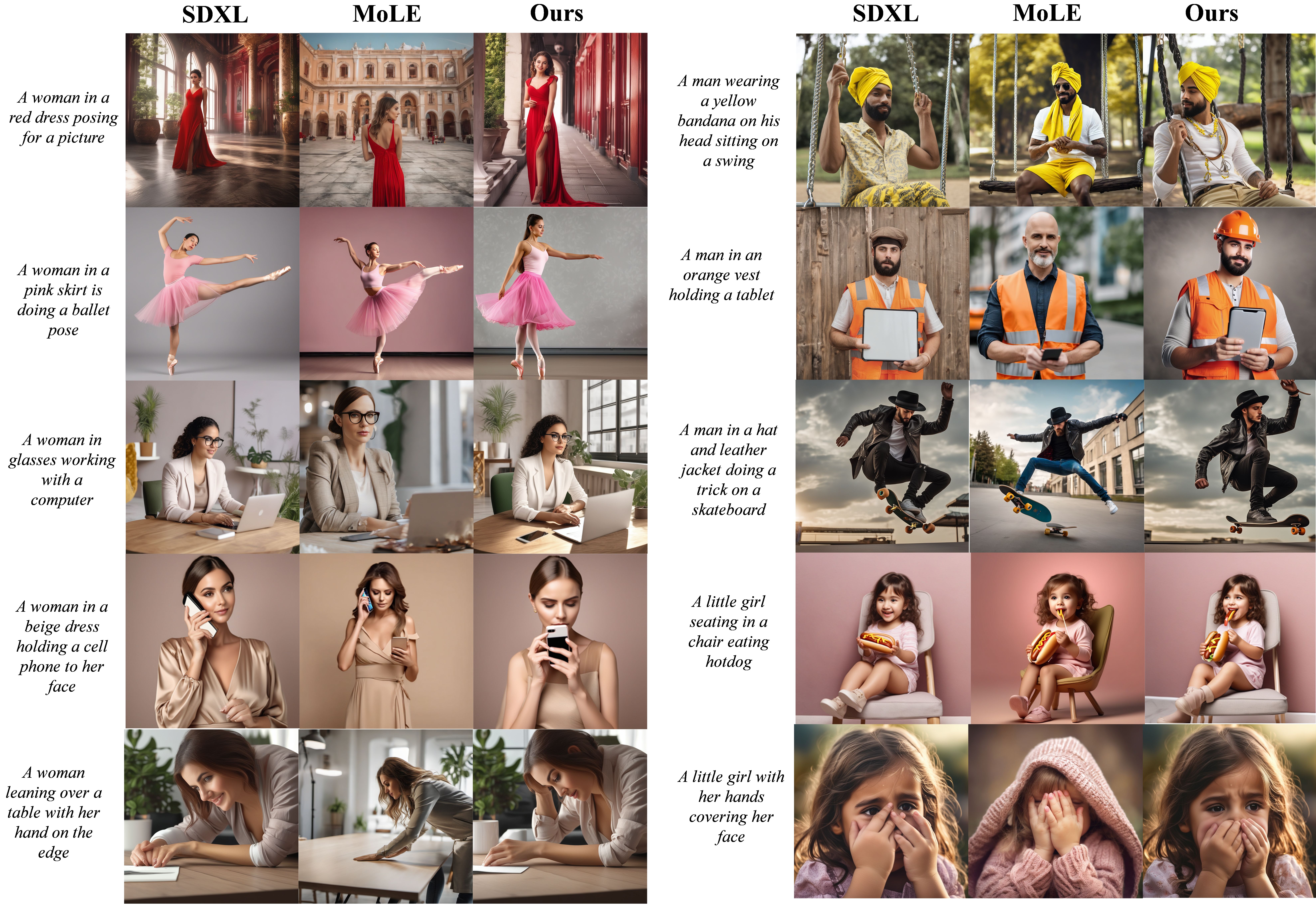}
\caption{Comparison with general T2I methods.}
\label{fig:s3}
\end{figure*}

\begin{figure*}[htb]
\centering
\includegraphics[width=\textwidth]{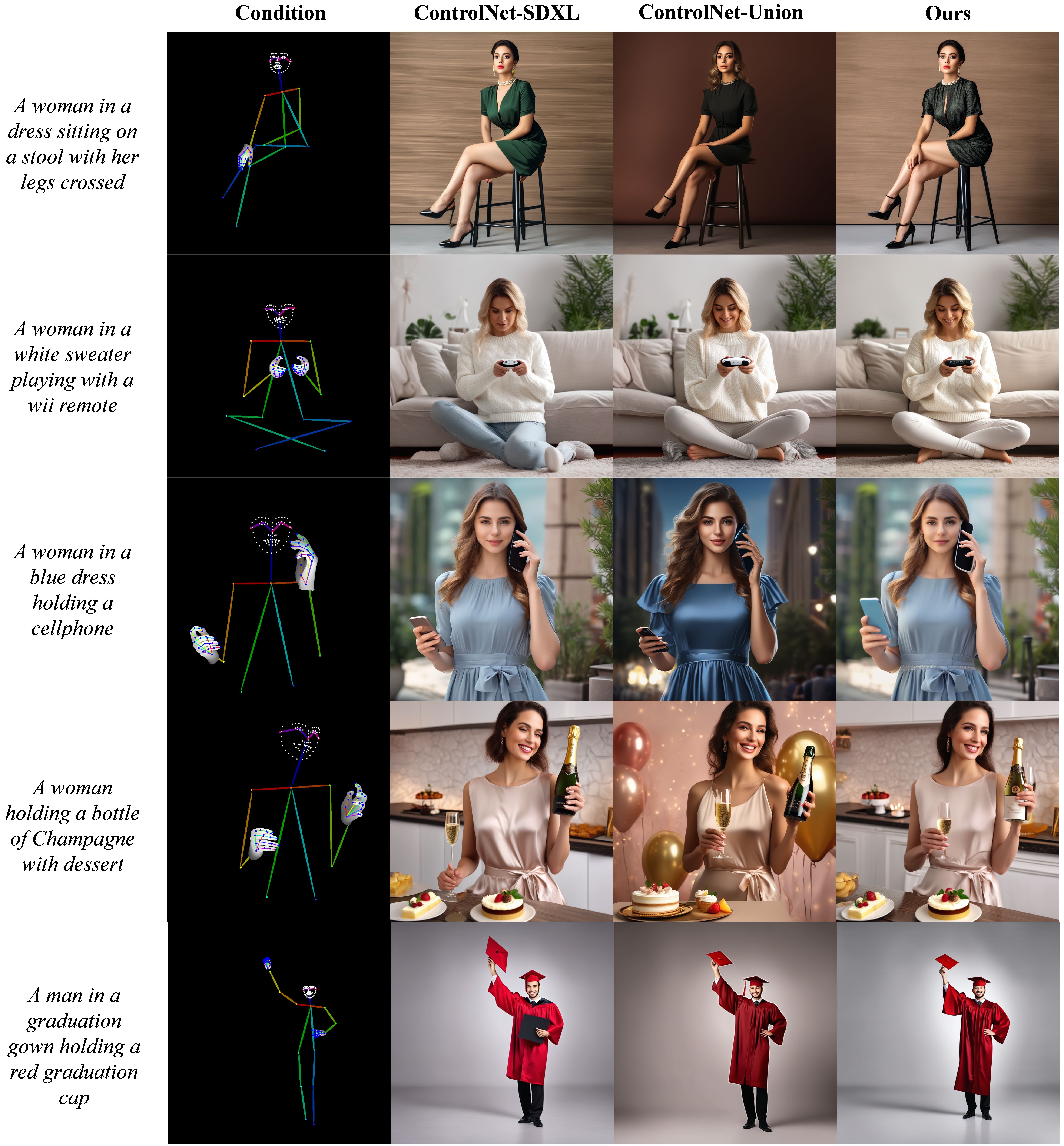}
\caption{Comparison with controllable methods.}
\label{fig:s4}
\end{figure*}

More visualization results compared to existing methods are demonstrated in Fig.\ref{fig:s1}, Fig.\ref{fig:s2}, Fig.\ref{fig:s3}, and Fig.\ref{fig:s4}. In addition to the baseline methods mentioned in our paper, we also compare two close-source methods for human image generation called HanDiffuser\cite{narasimhaswamy2024handiffuser} and HyperHuman\cite{liu2023hyperhuman}. Here, we extract example images from their paper for comparison.

%% file: main.bbl
\begin{thebibliography}{52}
\providecommand{\natexlab}[1]{#1}
\providecommand{\url}[1]{\texttt{#1}}
\expandafter\ifx\csname urlstyle\endcsname\relax
  \providecommand{\doi}[1]{doi: #1}\else
  \providecommand{\doi}{doi: \begingroup \urlstyle{rm}\Url}\fi

\bibitem[Ban and Ji(2024)]{ban2024fair}
Hao Ban and Kaiyi Ji.
\newblock Fair resource allocation in multi-task learning.
\newblock \emph{arXiv preprint arXiv:2402.15638}, 2024.

\bibitem[Betker et~al.(2023)Betker, Goh, Jing, Brooks, Wang, Li, Ouyang, Zhuang, Lee, Guo, et~al.]{betker2023improving}
James Betker, Gabriel Goh, Li Jing, Tim Brooks, Jianfeng Wang, Linjie Li, Long Ouyang, Juntang Zhuang, Joyce Lee, Yufei Guo, et~al.
\newblock Improving image generation with better captions.
\newblock \emph{Computer Science. https://cdn. openai. com/papers/dall-e-3. pdf}, 2\penalty0 (3):\penalty0 8, 2023.

\bibitem[Bing-su(2024)]{adetailer}
Bing-su.
\newblock Adetailer.
\newblock \url{https://github.com/Bing-su/adetailer.git}, 2024.

\bibitem[Choi et~al.(2021)Choi, Park, Lee, and Choo]{choi2021viton}
Seunghwan Choi, Sunghyun Park, Minsoo Lee, and Jaegul Choo.
\newblock Viton-hd: High-resolution virtual try-on via misalignment-aware normalization.
\newblock In \emph{Proceedings of the IEEE/CVF conference on computer vision and pattern recognition}, pages 14131--14140, 2021.

\bibitem[Esser et~al.(2024)Esser, Kulal, Blattmann, Entezari, M{\"u}ller, Saini, Levi, Lorenz, Sauer, Boesel, et~al.]{esser2024scaling}
Patrick Esser, Sumith Kulal, Andreas Blattmann, Rahim Entezari, Jonas M{\"u}ller, Harry Saini, Yam Levi, Dominik Lorenz, Axel Sauer, Frederic Boesel, et~al.
\newblock Scaling rectified flow transformers for high-resolution image synthesis.
\newblock In \emph{Forty-first International Conference on Machine Learning}, 2024.

\bibitem[Fu et~al.(2022)Fu, Li, Jiang, Lin, Qian, Loy, Wu, and Liu]{fu2022stylegan}
Jianglin Fu, Shikai Li, Yuming Jiang, Kwan-Yee Lin, Chen Qian, Chen~Change Loy, Wayne Wu, and Ziwei Liu.
\newblock Stylegan-human: A data-centric odyssey of human generation.
\newblock In \emph{European Conference on Computer Vision}, pages 1--19. Springer, 2022.

\bibitem[Gao et~al.(2024)Gao, Chen, Cui, Ren, Wang, Zhu, Tian, Ye, He, Zhu, et~al.]{gao2024mini}
Zhangwei Gao, Zhe Chen, Erfei Cui, Yiming Ren, Weiyun Wang, Jinguo Zhu, Hao Tian, Shenglong Ye, Junjun He, Xizhou Zhu, et~al.
\newblock Mini-internvl: a flexible-transfer pocket multi-modal model with 5\% parameters and 90\% performance.
\newblock \emph{Visual Intelligence}, 2\penalty0 (1):\penalty0 1--17, 2024.

\bibitem[Heusel et~al.(2017)Heusel, Ramsauer, Unterthiner, Nessler, and Hochreiter]{heusel2017gans}
Martin Heusel, Hubert Ramsauer, Thomas Unterthiner, Bernhard Nessler, and Sepp Hochreiter.
\newblock Gans trained by a two time-scale update rule converge to a local nash equilibrium.
\newblock \emph{Advances in neural information processing systems}, 30, 2017.

\bibitem[Ho et~al.(2020)Ho, Jain, and Abbeel]{ho2020denoising}
Jonathan Ho, Ajay Jain, and Pieter Abbeel.
\newblock Denoising diffusion probabilistic models.
\newblock \emph{Advances in neural information processing systems}, 33:\penalty0 6840--6851, 2020.

\bibitem[Hu et~al.(2021)Hu, Shen, Wallis, Allen-Zhu, Li, Wang, Wang, and Chen]{hu2021lora}
Edward~J Hu, Yelong Shen, Phillip Wallis, Zeyuan Allen-Zhu, Yuanzhi Li, Shean Wang, Lu Wang, and Weizhu Chen.
\newblock Lora: Low-rank adaptation of large language models.
\newblock \emph{arXiv preprint arXiv:2106.09685}, 2021.

\bibitem[Hu et~al.(2024)Hu, Tu, Han, He, Cui, Long, Zheng, Fang, Huang, Zhao, et~al.]{hu2024minicpm}
Shengding Hu, Yuge Tu, Xu Han, Chaoqun He, Ganqu Cui, Xiang Long, Zhi Zheng, Yewei Fang, Yuxiang Huang, Weilin Zhao, et~al.
\newblock Minicpm: Unveiling the potential of small language models with scalable training strategies.
\newblock \emph{arXiv preprint arXiv:2404.06395}, 2024.

\bibitem[Ionescu et~al.(2013)Ionescu, Papava, Olaru, and Sminchisescu]{ionescu2013human3}
Catalin Ionescu, Dragos Papava, Vlad Olaru, and Cristian Sminchisescu.
\newblock Human3. 6m: Large scale datasets and predictive methods for 3d human sensing in natural environments.
\newblock \emph{IEEE transactions on pattern analysis and machine intelligence}, 36\penalty0 (7):\penalty0 1325--1339, 2013.

\bibitem[Jacobs et~al.(1991)Jacobs, Jordan, Nowlan, and Hinton]{jacobs1991adaptive}
Robert~A Jacobs, Michael~I Jordan, Steven~J Nowlan, and Geoffrey~E Hinton.
\newblock Adaptive mixtures of local experts.
\newblock \emph{Neural computation}, 3\penalty0 (1):\penalty0 79--87, 1991.

\bibitem[Jia et~al.(2024)Jia, Zhang, Wang, and Tan]{jia2024human}
Zhen Jia, Zhang Zhang, Liang Wang, and Tieniu Tan.
\newblock Human image generation: A comprehensive survey.
\newblock \emph{ACM Computing Surveys}, 56\penalty0 (11):\penalty0 1--39, 2024.

\bibitem[Ju et~al.(2023{\natexlab{a}})Ju, Zeng, Wang, Xu, and Zhang]{ju2023human}
Xuan Ju, Ailing Zeng, Jianan Wang, Qiang Xu, and Lei Zhang.
\newblock Human-art: A versatile human-centric dataset bridging natural and artificial scenes.
\newblock In \emph{Proceedings of the IEEE/CVF Conference on Computer Vision and Pattern Recognition}, pages 618--629, 2023{\natexlab{a}}.

\bibitem[Ju et~al.(2023{\natexlab{b}})Ju, Zeng, Zhao, Wang, Zhang, and Xu]{ju2023humansd}
Xuan Ju, Ailing Zeng, Chenchen Zhao, Jianan Wang, Lei Zhang, and Qiang Xu.
\newblock Humansd: A native skeleton-guided diffusion model for human image generation.
\newblock In \emph{Proceedings of the IEEE/CVF International Conference on Computer Vision}, pages 15988--15998, 2023{\natexlab{b}}.

\bibitem[Kendall et~al.(2018)Kendall, Gal, and Cipolla]{kendall2018multi}
Alex Kendall, Yarin Gal, and Roberto Cipolla.
\newblock Multi-task learning using uncertainty to weigh losses for scene geometry and semantics.
\newblock In \emph{Proceedings of the IEEE conference on computer vision and pattern recognition}, pages 7482--7491, 2018.

\bibitem[Kingma(2013)]{kingma2013auto}
Diederik~P Kingma.
\newblock Auto-encoding variational bayes.
\newblock \emph{arXiv preprint arXiv:1312.6114}, 2013.

\bibitem[Li et~al.(2022)Li, Li, Xiong, and Hoi]{li2022blip}
Junnan Li, Dongxu Li, Caiming Xiong, and Steven Hoi.
\newblock Blip: Bootstrapping language-image pre-training for unified vision-language understanding and generation.
\newblock In \emph{International conference on machine learning}, pages 12888--12900. PMLR, 2022.

\bibitem[Li et~al.(2024{\natexlab{a}})Li, Yang, Kuang, Wu, Wang, Xiao, and Chen]{li2024controlnet++}
Ming Li, Taojiannan Yang, Huafeng Kuang, Jie Wu, Zhaoning Wang, Xuefeng Xiao, and Chen Chen.
\newblock Controlnet++: Improving conditional controls with efficient consistency feedback: Project page: liming-ai. github. io/controlnet\_plus\_plus.
\newblock In \emph{European Conference on Computer Vision}, pages 129--147. Springer, 2024{\natexlab{a}}.

\bibitem[Li et~al.(2024{\natexlab{b}})Li, Fu, Liu, Wang, Lin, and Wu]{li2024cosmicman}
Shikai Li, Jianglin Fu, Kaiyuan Liu, Wentao Wang, Kwan-Yee Lin, and Wayne Wu.
\newblock Cosmicman: A text-to-image foundation model for humans.
\newblock In \emph{Proceedings of the IEEE/CVF Conference on Computer Vision and Pattern Recognition}, pages 6955--6965, 2024{\natexlab{b}}.

\bibitem[Lin et~al.(2021)Lin, Ye, Zhang, and Tsang]{lin2021reasonable}
Baijiong Lin, Feiyang Ye, Yu Zhang, and Ivor~W Tsang.
\newblock Reasonable effectiveness of random weighting: A litmus test for multi-task learning.
\newblock \emph{arXiv preprint arXiv:2111.10603}, 2021.

\bibitem[Liu et~al.(2019)Liu, Johns, and Davison]{liu2019end}
Shikun Liu, Edward Johns, and Andrew~J Davison.
\newblock End-to-end multi-task learning with attention.
\newblock In \emph{Proceedings of the IEEE/CVF conference on computer vision and pattern recognition}, pages 1871--1880, 2019.

\bibitem[Liu et~al.(2023)Liu, Ren, Siarohin, Skorokhodov, Li, Lin, Liu, Liu, and Tulyakov]{liu2023hyperhuman}
Xian Liu, Jian Ren, Aliaksandr Siarohin, Ivan Skorokhodov, Yanyu Li, Dahua Lin, Xihui Liu, Ziwei Liu, and Sergey Tulyakov.
\newblock Hyperhuman: Hyper-realistic human generation with latent structural diffusion.
\newblock \emph{arXiv preprint arXiv:2310.08579}, 2023.

\bibitem[Lu et~al.(2024)Lu, Xu, Zhang, Wang, and Tao]{lu2024handrefiner}
Wenquan Lu, Yufei Xu, Jing Zhang, Chaoyue Wang, and Dacheng Tao.
\newblock Handrefiner: Refining malformed hands in generated images by diffusion-based conditional inpainting.
\newblock In \emph{Proceedings of the 32nd ACM International Conference on Multimedia}, pages 7085--7093, 2024.

\bibitem[Lugaresi et~al.(2019)Lugaresi, Tang, Nash, McClanahan, Uboweja, Hays, Zhang, Chang, Yong, Lee, et~al.]{lugaresi2019mediapipe}
Camillo Lugaresi, Jiuqiang Tang, Hadon Nash, Chris McClanahan, Esha Uboweja, Michael Hays, Fan Zhang, Chuo-Ling Chang, Ming~Guang Yong, Juhyun Lee, et~al.
\newblock Mediapipe: A framework for building perception pipelines.
\newblock \emph{arXiv preprint arXiv:1906.08172}, 2019.

\bibitem[Massouli{\'e} and Roberts(1999)]{massoulie1999bandwidth}
Laurent Massouli{\'e} and James Roberts.
\newblock Bandwidth sharing: objectives and algorithms.
\newblock In \emph{IEEE INFOCOM'99. Conference on Computer Communications. Proceedings. Eighteenth Annual Joint Conference of the IEEE Computer and Communications Societies. The Future is Now (Cat. No. 99CH36320)}, pages 1395--1403. IEEE, 1999.

\bibitem[Mo and Walrand(2000)]{mo2000fair}
Jeonghoon Mo and Jean Walrand.
\newblock Fair end-to-end window-based congestion control.
\newblock \emph{IEEE/ACM Transactions on networking}, 8\penalty0 (5):\penalty0 556--567, 2000.

\bibitem[Narasimhaswamy et~al.(2022)Narasimhaswamy, Nguyen, Huang, and Hoai]{narasimhaswamy2022whose}
Supreeth Narasimhaswamy, Thanh Nguyen, Mingzhen Huang, and Minh Hoai.
\newblock Whose hands are these? hand detection and hand-body association in the wild.
\newblock In \emph{Proceedings of the IEEE/CVF Conference on Computer Vision and Pattern Recognition}, pages 4889--4899, 2022.

\bibitem[Narasimhaswamy et~al.(2024)Narasimhaswamy, Bhattacharya, Chen, Dasgupta, Mitra, and Hoai]{narasimhaswamy2024handiffuser}
Supreeth Narasimhaswamy, Uttaran Bhattacharya, Xiang Chen, Ishita Dasgupta, Saayan Mitra, and Minh Hoai.
\newblock Handiffuser: Text-to-image generation with realistic hand appearances.
\newblock In \emph{Proceedings of the IEEE/CVF Conference on Computer Vision and Pattern Recognition}, pages 2468--2479, 2024.

\bibitem[Navon et~al.(2022)Navon, Shamsian, Achituve, Maron, Kawaguchi, Chechik, and Fetaya]{navon2022multi}
Aviv Navon, Aviv Shamsian, Idan Achituve, Haggai Maron, Kenji Kawaguchi, Gal Chechik, and Ethan Fetaya.
\newblock Multi-task learning as a bargaining game.
\newblock \emph{arXiv preprint arXiv:2202.01017}, 2022.

\bibitem[Ngatchou et~al.(2005)Ngatchou, Zarei, and El-Sharkawi]{ngatchou2005pareto}
Patrick Ngatchou, Anahita Zarei, and A El-Sharkawi.
\newblock Pareto multi objective optimization.
\newblock In \emph{Proceedings of the 13th international conference on, intelligent systems application to power systems}, pages 84--91. IEEE, 2005.

\bibitem[Pavlakos et~al.(2024)Pavlakos, Shan, Radosavovic, Kanazawa, Fouhey, and Malik]{pavlakos2024reconstructing}
Georgios Pavlakos, Dandan Shan, Ilija Radosavovic, Angjoo Kanazawa, David Fouhey, and Jitendra Malik.
\newblock Reconstructing hands in 3d with transformers.
\newblock In \emph{Proceedings of the IEEE/CVF Conference on Computer Vision and Pattern Recognition}, pages 9826--9836, 2024.

\bibitem[Podell et~al.(2023)Podell, English, Lacey, Blattmann, Dockhorn, M{\"u}ller, Penna, and Rombach]{podell2023sdxl}
Dustin Podell, Zion English, Kyle Lacey, Andreas Blattmann, Tim Dockhorn, Jonas M{\"u}ller, Joe Penna, and Robin Rombach.
\newblock Sdxl: Improving latent diffusion models for high-resolution image synthesis.
\newblock \emph{arXiv preprint arXiv:2307.01952}, 2023.

\bibitem[Puigcerver et~al.(2023)Puigcerver, Riquelme, Mustafa, and Houlsby]{puigcerver2023sparse}
Joan Puigcerver, Carlos Riquelme, Basil Mustafa, and Neil Houlsby.
\newblock From sparse to soft mixtures of experts.
\newblock \emph{arXiv preprint arXiv:2308.00951}, 2023.

\bibitem[Radford et~al.(2021)Radford, Kim, Hallacy, Ramesh, Goh, Agarwal, Sastry, Askell, Mishkin, Clark, et~al.]{radford2021learning}
Alec Radford, Jong~Wook Kim, Chris Hallacy, Aditya Ramesh, Gabriel Goh, Sandhini Agarwal, Girish Sastry, Amanda Askell, Pamela Mishkin, Jack Clark, et~al.
\newblock Learning transferable visual models from natural language supervision.
\newblock In \emph{International conference on machine learning}, pages 8748--8763. PmLR, 2021.

\bibitem[Ramesh et~al.(2022)Ramesh, Dhariwal, Nichol, Chu, and Chen]{ramesh2022hierarchical}
Aditya Ramesh, Prafulla Dhariwal, Alex Nichol, Casey Chu, and Mark Chen.
\newblock Hierarchical text-conditional image generation with clip latents.
\newblock \emph{arXiv preprint arXiv:2204.06125}, 1\penalty0 (2):\penalty0 3, 2022.

\bibitem[Redmon(2016)]{redmon2016you}
J Redmon.
\newblock You only look once: Unified, real-time object detection.
\newblock In \emph{Proceedings of the IEEE conference on computer vision and pattern recognition}, 2016.

\bibitem[Rombach et~al.(2022)Rombach, Blattmann, Lorenz, Esser, and Ommer]{rombach2022high}
Robin Rombach, Andreas Blattmann, Dominik Lorenz, Patrick Esser, and Bj{\"o}rn Ommer.
\newblock High-resolution image synthesis with latent diffusion models.
\newblock In \emph{Proceedings of the IEEE/CVF conference on computer vision and pattern recognition}, pages 10684--10695, 2022.

\bibitem[Sauer et~al.(2023)Sauer, Karras, Laine, Geiger, and Aila]{sauer2023stylegan}
Axel Sauer, Tero Karras, Samuli Laine, Andreas Geiger, and Timo Aila.
\newblock Stylegan-t: Unlocking the power of gans for fast large-scale text-to-image synthesis.
\newblock In \emph{International conference on machine learning}, pages 30105--30118. PMLR, 2023.

\bibitem[Sener and Koltun(2018)]{sener2018multi}
Ozan Sener and Vladlen Koltun.
\newblock Multi-task learning as multi-objective optimization.
\newblock \emph{Advances in neural information processing systems}, 31, 2018.

\bibitem[Song et~al.(2020)Song, Meng, and Ermon]{song2020denoising}
Jiaming Song, Chenlin Meng, and Stefano Ermon.
\newblock Denoising diffusion implicit models.
\newblock \emph{arXiv preprint arXiv:2010.02502}, 2020.

\bibitem[Wang et~al.(2024)Wang, Cao, Liang, He, Sun, Li, and Ma]{wang2024mixture}
Yuxuan Wang, Tianwei Cao, Kongming Liang, Zhongjiang He, Hao Sun, Yongxiang Li, and Zhanyu Ma.
\newblock Mixture-of-hand-experts: Repainting the deformed hand images generated by diffusion models.
\newblock In \emph{Chinese Conference on Pattern Recognition and Computer Vision (PRCV)}, pages 143--157. Springer, 2024.

\bibitem[Wu et~al.(2023)Wu, Hao, Sun, Chen, Zhu, Zhao, and Li]{wu2023human}
Xiaoshi Wu, Yiming Hao, Keqiang Sun, Yixiong Chen, Feng Zhu, Rui Zhao, and Hongsheng Li.
\newblock Human preference score v2: A solid benchmark for evaluating human preferences of text-to-image synthesis.
\newblock \emph{arXiv preprint arXiv:2306.09341}, 2023.

\bibitem[Xu et~al.(2023)Xu, Liu, Wu, Tong, Li, Ding, Tang, and Dong]{xu2023imagereward}
Jiazheng Xu, Xiao Liu, Yuchen Wu, Yuxuan Tong, Qinkai Li, Ming Ding, Jie Tang, and Yuxiao Dong.
\newblock Imagereward: Learning and evaluating human preferences for text-to-image generation.
\newblock \emph{Advances in Neural Information Processing Systems}, 36:\penalty0 15903--15935, 2023.

\bibitem[Yang et~al.(2023)Yang, Zeng, Yuan, and Li]{yang2023effective}
Zhendong Yang, Ailing Zeng, Chun Yuan, and Yu Li.
\newblock Effective whole-body pose estimation with two-stages distillation.
\newblock In \emph{Proceedings of the IEEE/CVF International Conference on Computer Vision}, pages 4210--4220, 2023.

\bibitem[Yu et~al.(2020)Yu, Kumar, Gupta, Levine, Hausman, and Finn]{yu2020gradient}
Tianhe Yu, Saurabh Kumar, Abhishek Gupta, Sergey Levine, Karol Hausman, and Chelsea Finn.
\newblock Gradient surgery for multi-task learning.
\newblock In \emph{Advances in Neural Information Processing Systems}, pages 5824--5836, 2020.

\bibitem[Zhang et~al.(2017)Zhang, Xu, Li, Zhang, Wang, Huang, and Metaxas]{zhang2017stackgan}
Han Zhang, Tao Xu, Hongsheng Li, Shaoting Zhang, Xiaogang Wang, Xiaolei Huang, and Dimitris~N Metaxas.
\newblock Stackgan: Text to photo-realistic image synthesis with stacked generative adversarial networks.
\newblock In \emph{Proceedings of the IEEE international conference on computer vision}, pages 5907--5915, 2017.

\bibitem[Zhang et~al.(2018)Zhang, Xu, Li, Zhang, Wang, Huang, and Metaxas]{zhang2018stackgan++}
Han Zhang, Tao Xu, Hongsheng Li, Shaoting Zhang, Xiaogang Wang, Xiaolei Huang, and Dimitris~N Metaxas.
\newblock Stackgan++: Realistic image synthesis with stacked generative adversarial networks.
\newblock \emph{IEEE transactions on pattern analysis and machine intelligence}, 41\penalty0 (8):\penalty0 1947--1962, 2018.

\bibitem[Zhang et~al.(2023)Zhang, Rao, and Agrawala]{zhang2023adding}
Lvmin Zhang, Anyi Rao, and Maneesh Agrawala.
\newblock Adding conditional control to text-to-image diffusion models.
\newblock In \emph{Proceedings of the IEEE/CVF International Conference on Computer Vision}, pages 3836--3847, 2023.

\bibitem[Zheng et~al.(2015)Zheng, Shen, Tian, Wang, Wang, and Tian]{zheng2015scalable}
Liang Zheng, Liyue Shen, Lu Tian, Shengjin Wang, Jingdong Wang, and Qi Tian.
\newblock Scalable person re-identification: A benchmark.
\newblock In \emph{Proceedings of the IEEE international conference on computer vision}, pages 1116--1124, 2015.

\bibitem[Zhu et~al.(2024)Zhu, Chen, Ding, Luo, Wang, and Wang]{zhu2024mole}
Jie Zhu, Yixiong Chen, Mingyu Ding, Ping Luo, Leye Wang, and Jingdong Wang.
\newblock Mole: Enhancing human-centric text-to-image diffusion via mixture of low-rank experts.
\newblock \emph{arXiv preprint arXiv:2410.23332}, 2024.

\end{thebibliography}
